\documentclass{article}

\usepackage{PRIMEarxiv}

\usepackage[utf8]{inputenc} 
\usepackage[T1]{fontenc}    
\usepackage{hyperref}       
\usepackage{url}            
\usepackage{booktabs}       
\usepackage{amsfonts}       
\usepackage{nicefrac}       
\usepackage{microtype}      
\usepackage{lipsum}
\usepackage{fancyhdr}       
\usepackage{graphicx}       
\graphicspath{{media/}}     

\usepackage[linesnumbered,ruled,vlined]{algorithm2e}
\usepackage{algorithmic}

\usepackage{subfigure}
\usepackage{amsmath}
\usepackage{amssymb}
\usepackage{amsthm}
\newtheorem{proposition}{Proposition}

\title{The Art of Manipulation:\\ Threat of Multi-Step Manipulative Attacks in Security Games}

\author{
  Thanh H. Nguyen \\
  University of Oregon \\
  \texttt{\{thanhhng\}@cs.uoregon.edu} \\
   \And
  Arunesh Sinha \\
  Singapore Management University \\
  \texttt{\{aruneshs\}@smu.edu.sg} \\
}
\begin{document}

\maketitle


\begin{abstract}
This paper studies the problem of multi-step manipulative attacks in Stackelberg security games, in which a clever attacker attempts to orchestrate its attacks over multiple time steps to mislead the defender's learning of the attacker's behavior. This attack manipulation eventually influences the defender's patrol strategy towards the attacker's benefit. Previous work along this line of research only focuses on one-shot games in which the defender learns the attacker's behavior and then designs a corresponding strategy only once. Our work, on the other hand, investigates the long-term impact of the attacker's manipulation in which current attack and defense choices of players determine the future learning and patrol planning of the defender. This paper has three key contributions. First, we introduce a new multi-step manipulative attack game model that captures the impact of sequential manipulative attacks carried out by the attacker over the entire time horizon. Second, we propose a new algorithm to compute an optimal manipulative attack plan for the attacker, which tackles the challenge of multiple connected optimization components involved in the computation across multiple time steps. Finally, we present extensive experimental results on the impact of such misleading attacks, showing a significant benefit for the attacker and loss for the defender.  
\end{abstract}





         







\section{Introduction}
Stackelberg security games (SSGs) have been widely applied for solving many real-world problems in public safety and security, cybersecurity, and conversations~\cite{pita2008deployed,tambe2011security,shieh12,fang2016deploying}. In fact, there have been several applications of SSGs such as ARMOR (used for protecting airport terminals at Los Angeles International Airport)~\cite{pita2008deployed}, PROTECT (used by US Coast Guard officers to protect ferries)~\cite{shieh12}, and PAWS (used by NGOs in multiple national parks across the world for protecting wildlife)~\cite{fang2016deploying}, etc. 
In recent work in SSGs, machine learning-based techniques have been used for modeling and predicting the attacker's behavior based on collected attack data~\cite{yang11,kar2017cloudy,sinha2018stackelberg,gholami19}.
For example, in PROTECT, Quantal Response was used to predict decision making of the attacker in the domain of ferry protection~\cite{shieh12}. In addition, in the PAWS-related work, different models such as Quantal Response, SUQR, and SHARP, etc were used to capture the behavior of poachers (i.e., predicting where the poachers are likely to set trapping tools to catch wild animals)~\cite{fang2016deploying,kar2017cloudy,kar2015game,sinha2018stackelberg,gholami19}. These behavior models, after being trained, are used to determine an optimal strategy of the defender. 

However, as pointed out in previous work, since the defender relies on some attack data to make prediction, the attacker can intentionally change its attack behavior to mislead the defender's learning~\cite{nguyendecoding,thanh2020}. Consequently, the learned adversary behavior model deviates from the true behavior model, and causes the defender to generate ineffective patrolling strategies, which benefits the attacker in the end. Intuitively, the attacker is perfectly rational, but pretends to act in a boundedly rational manner. The attacker may suffer an immediate loss for deviating from a myopic optimal response, but it will gain significantly more long-term benefit as a result of the defender's deteriorated strategies. In this work, we focus on analyzing such manipulative attacks of the attacker. 

The existing works in SSGs mainly study one-shot game scenarios in which the defender only learns the attacker behavior once and then commits to a single defense strategy afterward~\cite{nguyendecoding,thanh2020}. However, in many real-world domains such as wildlife protection, the defender and attacker often interact in a repeated manner~\cite{fang2016deploying}. That is, at each time step, given historical attack and patrol data, the defender updates his model of the attacker's behavior and re-generates a new defense strategy while the attacker responds accordingly by launching a certain number of attacks. These new defense and attack actions are then collected for the future use. This learning-patrolling-attack loop continues until the end of the time horizon. In this multi-step interaction scenario, it is clear that the existing one-shot SSG studies fail to capture the long-term impact of the attacker's manipulation. 

In this work, we study the problem of sequential manipulative attacks in multi-step SSGs. We aim at investigating the long-term manipulative decisions of the attacker and the accumulative impact of such manipulation on both the defender and attacker's utility. We provide the following three key contributions. First, we introduce a new multi-step manipulative attack game model. In our game model, the defender follows a learning-patrolling process at each time step to play. On the other hand, at each time step, the attacker attempts to find an optimal attack strategy given the current defense strategy, taking into account the tradeoff between the immediate utility loss for playing boundedly rational at current time step and the future utility gain for misleading the defender. Second, we present a new algorithm to compute such optimal manipulative attack plan for the attacker. The key challenge of computing an optimal attack plan is that it involves multiple connected optimization components over the entire time horizon, which is not straightforward to solve.  In order to tackle this computational challenge, our new algorithm follows the Projected Gradient Descent (PGD) approach to iteratively update the attack plan based on the gradient of the attacker's utility with respect to its attacks.  Inspired by hyper-parameter learning~\cite{bengio2000gradient}, we then determine this gradient based on the recursive relationships of the gradient components involved in the gradient updating steps of the inner optimization levels. 

Finally, we provide an extensive experimental analysis on the impact of the attacker's attack manipulation on the accumulated utility of both players. We show that the attacker gains a substantially higher utility while the defender suffers a significant loss as a result of the attacker's manipulation.   
\section{Related Work}
Attacker behavior modeling is an important research line in SSGs which focuses on building behavior models of the attacker in various security-related domains such as wildlife protection~\cite{yang11,kar2017cloudy,gholami19}. Several different models were proposed before, including Quantal Response~\cite{yang11}, SUQR~\cite{nguyen2013analyzing}, and SHARP~\cite{kar2015game} models. These models enable the defender to predict boundedly rational decisions of human attackers such as poachers using historical attack data as well as other domain-dependent feature data. For example, in wildlife protection, rangers can collect poaching signs such as snares during their patrols~\cite{fang2016deploying}. These observations are then used to predict poaching activities in the future. 

However, there is a rising concern about the vulnerability of these learning-patrolling methods in the presence of a deceptive attacker who intentionally maneuvers its attacks to \emph{fool} the defender's learning. 
Previous work has demonstrated that weakness of the learning-patrolling methods in one-shot (sometimes Bayesian) SSGs~\cite{gan2019imitative,nguyendecoding,thanh2020}; our focus is on multiple steps.
A multi-step related work has so far looked into a simple learning situation in which the defender uses the Bayes rule method to update his belief about the attacker's type over time~\cite{nguyen2019deception}. 

Our work is also related to adversarial learning in machine learning in the sense of attacking the training/testing data or interfering with the learning process. Poisoning attacks (i.e., altering the training data) are the most closely related to our work~\cite{lowd2005adversarial,Song18,huang2011,aleks2017deep,zhang2019online,demontis2019adversarial,biggio2018wild,papernot2018sok}. Different attack methods were designed to deteriorate the performance of standard machine learning algorithms such as SVMs and neural nets, etc. Differentiating from this research line, in our problem, multi-step decision quality (which is measured via utilities of players) in terms of players' strategies is the ultimate objective of the players, rather than just the prediction accuracy. 

Finally, in secrecy and deception in SSGs, previous work investigated situations in which information available to the defender and attacker is asymmetric~\cite{guo2017comparing,xu2015exploring,rabinovich2015information,hendricks2006feints,brown2005two,farrell1996cheap,Zhuang10}. They then determine how the defender should strategically reveal or disguise his information to the attacker so as to influence the attacker's decision for the sake of the defender's benefit.  

\section{Preliminaries}
\paragraph{Stackelberg security games (SSGs).}SSGs are a class of leader-follower games in which a defender has to allocate a limited number of security resources over a set of important targets $[N] = \{1, \dots, N\}$ to protect these targets against an attacker. 
In one-shot SSGs, a pure strategy of the defender can be viewed as a \emph{subset} of targets that can be covered by his security resources. A mixed strategy of the defender is a distribution over the defender's pure strategies. We consider generic SSGs in which the defender's mixed strategies can be represented as a marginal probability vector $\mathbf{x} = \{x_1,\dots, x_N\}$ with resource constraints can be captured by a set of linear constraints $\mathbf{A}\mathbf{x}\leq b$. Here, $x_n\in [0,1]$ is the marginal coverage of the defender at target $n$. We denote by $\mathbf{X} = \{\mathbf{x}: \mathbf{A}\mathbf{x}\leq b\}$ the set of all mixed strategies.

In SSGs, the players' payoff depends on which target the attacker attacks and whether the defender is protecting that target or not. In particular, when the attacker attacks a target $n$, if the defender is not protecting $n$, the attacker will receive a reward of $R^a_n$ while the defender gets a penalty of $P^a_n$. Conversely, if the defender is protecting $n$, the attacker gets a penalty $P^a_n < R^a_n$ and the defender obtains a reward $R^d_n > P^d_n$. Given a mixed strategy of the defender $\mathbf{x}$, when the attacker attacks $n$, the defender and attacker's expected utility at $n$ is computed as follows:
\begin{align}
    & U^d_n(x_n) = x_n(R^d_n - P^d_n) + P^d_n \\
    & U^a_n(x_n) = x_n(P^a_n - R^a_n) + R^a_n
\end{align}
A standard game-theoretic solution concept in SSGs is Strong Stackelberg Equilibrium (SSE) in which players play optimally against each other given that the attacker is aware of the defender strategy before taking any action. 
    

\paragraph{Attacker behavior models.}One of the very first behavior model used to predict the attacker behavior is Quantal Response, a well-known behavior model used in both behavioral economics and game theory~~\cite{mcfadden1973conditional,mckelvey1995quantal,yang11}. While an SSE considers a perfectly rational attacker, QR assumes a boundedly rational attacker who attacks each target $n$ with a probability proportional to the attacker's expected utility at that target. Later on, SUQR, an extension of Quantal Response, which uses a linear combination of various domain features to reason about the attacker's behavior~\cite{nguyen2013analyzing}. Building upon the success of QR and SUQR, Kar et. al introduce a new model, named SHARP, that augments a two-parameter probability weighting function into the SUQR model to predict poachers' behavior in wildlife protection~\cite{kar2015game}. 
Among all these models, the final prediction of the attacker's behavior can be abstractly captured using the following soft-max function:
\begin{align*}
    & q_n(\mathbf{x}, \theta) = \frac{e^{f(x_n,\theta)}}{\sum_{n'} e^{f(x_{n'},\theta)}}
\end{align*}
which is the probability the attacker will attacks target $n$. The function $f({x}_n,\theta)$ represents the behavior model used by the defender, indicating that the attacker's decision depends on the defender's strategy ${x}_n$ (in addition to other domain features such as the attacker rewards and penalties, etc. which we omit from the formulation for the sake of presentation). Finally, $\theta \in \mathbb{R}^m$ is the model parameter vector. 
\section{Manipulative Attack Game Model}
In many real-world security domains such as wildlife protection, the defender and attacker repeatedly interact with each other through a multi-step learning-patrolling-attacking loop. The one-shot SSG model can be then extended to capture such security scenarios. 

Formally, at each time step $t$, let's denote by $(\mathbf{X}_{t-1}, \mathbf{Z}_{t-1})$ the historical patrolling strategies and attacks at previous time steps. $(\mathbf{X}_{t-1}, \mathbf{Z}_{t-1})$ is also the data the defender uses to learn the attacker's behavior. In particular, $\mathbf{X}_{t-1} = \{\mathbf{x}_1,\dots, \mathbf{x}_{t-1}\}$ where $\mathbf{x}_{t'} = \{x_{t', 1},x_{t',2},\dots, x_{t',N}\}$ with $t'\leq t-1$ is the defender's mixed strategy at time step $t'$. In addition, $\mathbf{Z}_{t-1} = \{\mathbf{z}_1,\dots, \mathbf{z}_{t-1}\}$ where $\mathbf{z}_{t'} = \{z_{t',1},z_{t',2},\dots, z_{t',N }\}$ is the attack distribution at time step $t'\leq t-1$ (i.e., $z_{t',n}$ is the number of times the attackers attacks target $n$ in time step $t'$). The horizon is $T$ timesteps and we denote $[T] = \{1, \ldots, T\}$.
\paragraph{Defender's learning and patrolling.}
At each time step $t$, the defender's strategy $\mathbf{x}_t$ follows a two-stage learning-patrolling process to determine his strategy at $t$:
\begin{itemize}
    \item Learning: The defender optimizes the model parameter vector $\theta_t$ at step $t$ based on $(\mathbf{X}_{t-1}, \mathbf{Z}_{t-1})$, which is the result of the following minimization problem:
    \begin{align*}
        \min\nolimits_{\theta\in\Theta}\; &L(\mathbf{X}_{t-1}, \mathbf{Z}_{t-1},\theta)
    \end{align*}
    where $L(\mathbf{X}_{t-1}, \mathbf{Z}_{t-1},\theta)$ is the defender's loss function and $\Theta$ is the set of possible values of $\theta$.
    \item Patrolling: Given the learning outcome $\theta_{t}$, the defender finds an optimal strategy $\mathbf{x}_t$ accordingly, which is an optimal solution of the following optimization problem:
    \begin{align*}
        \max\nolimits_{\mathbf{x}\in\mathbf{X}}\;& U^d(\mathbf{x},\theta_t)
    \end{align*}
    which maximizes the defender's utility with respect to the learned parameter $\theta_t$, denoted by $U^d(\mathbf{x},\theta_t)$, with:
\begin{align*}
    U^d(\mathbf{x},\theta_t) = \sum\nolimits_n q_n(\mathbf{x},\theta_t) U^d_n(x_n)
\end{align*}
\end{itemize}
At the first time step $t = 1$, in particular, the defender does not have any data. Therefore, the defender can choose a particular strategy $\mathbf{x}_1$ to play, such as the $\mathtt{SSE}$ strategy. This repeated learning-patrolling process has been used in the PAWS application for generating ranger patrols in the wildlife protection domain~\cite{fang2016deploying}. 
\paragraph{Attacker Manipulation.}Since the defender relies on attack data to learn the attacker's behavior, a clever attacker can orchestrate attacks to \emph{fool} the defender, influencing the defender's learning and as a result, leading to ineffective patrolling strategies which benefit the attacker. In our model, the attacker is perfectly rational, but \emph{pretends} to be bounded rational to mislead the defender. By acting in this manipulative way, the attacker suffers some immediate utility loss (for playing bounded rational) but would obtain a significant long term benefit as the result of its influence on the defender's patrolling strategies. The attacker's goal is to find an optimal manipulative sequential-attack strategy that maximizes the attacker's accumulative expected utility across the entire time horizon, given the trade-off between the loss and benefit of such pretentious bounded rational playing. 

In this paper, we will focus on analyzing such manipulative attacks, assuming the attacker knows the defender's learning model. In real-world security domains, the attacker may not know the learning model used by the defender. In this case, the attacker can assume a certain behavior model and plan its deceptive attacks accordingly (this assumed model may be different from the behavior model used by the defender). Later in the experiment section, we will analyze the impact of the attacker's knowledge and model assumption on the attacker manipulation outcomes. 

Formally, the problem of finding an optimal manipulative attack strategy can be represented as the following: 
\begin{align}\label{dec.0}
    \max_{\mathbf{z}}\; &\sum\nolimits_t U^{a}(\mathbf{x}_t, \mathbf{z}_t)\\\label{dec.1}
    \text{s.t. }& \theta_t \in \arg\min\nolimits_{\theta\in\Theta} L(\mathbf{X}_{t-1}, \mathbf{Z}_{t-1}, \theta),\forall t\\\label{dec.2}
    & \mathbf{x}_t \in \arg\max\nolimits_{\mathbf{x}\in\mathbf{X}} U^d(\mathbf{x},\theta_t),\forall t\\\label{dec.3}
    & \sum\nolimits_n z_{t,n} \leq K,z_{t,n} \in \mathbb{N},\forall n, t
\end{align}
which maximizes the attacker's accumulated expected utility over the entire time horizon. In particular, the expected utility of the attacker at time step $t$ is computed as follows:
$$U^{a}(\mathbf{x}_t, \mathbf{z}_t) = \sum\nolimits_{n} z_{t,n} U^a_{n}(x_{t,n}) $$
Constraints~(\ref{dec.1}--\ref{dec.2}) represent the two-stage learning-patrolling of the defender at each time step $t$. Constraint (\ref{dec.3}) ensures that the attacker can only launch at most $K$ attacks at each time step. The constant $K$ represents the attacker's limited capability in influencing the defender's learning.

\section{Attack Manipulation Computation}

Overall, the problem (\ref{dec.0}--\ref{dec.3}) consists of multiple connected optimization levels. The decision on which targets and how frequently to attack at each time step not only influences the utility outcome at current time step but also affects the learning outcomes of the defender in future time steps. As a result, that attack decision of the attacker will contribute to the future utility outcomes that the attacker will receive. The problem (\ref{dec.0}--\ref{dec.3}) is challenging to solve. We propose to relax the attack variables $\{z_{t,n}\}$ to be continuous and then apply the Projected Gradient Descent (PGD) approach to solve it. Essentially, starting with some initial values of $\mathbf{z}^0=\{\mathbf{z}_{1}^0,\mathbf{z}^0_2,\dots, \mathbf{z}^0_T\}$, PGD iteratively updates the values of these attack variables based on the gradient step.   
Denote by $F = \sum_t U^{a}(\mathbf{x}_t, \mathbf{z}_t)$, at each iteration $i$ of the PGD, given the current estimation $\mathbf{z}^{i-1} = \{\mathbf{z}^{i-1}_1,\mathbf{z}^{i-1}_2,\dots, \mathbf{z}^{i-1}_T\}$, the gradient update step is as follows:
\begin{align}\label{PGD.1}
    &\mathbf{z}^i =  \mathbf{z}^{i-1} + \alpha \frac{dF}{d\mathbf{z}^{i-1}} 
\end{align} 
where $\alpha > 0$ is the step size. PGD then projects the updated value into the feasible region by finding the closest point in the region $\mathbf{Z} = \{\mathbf{z}: \sum_{n} z_{t,n} \leq K, z_{t,n} \geq 0, \forall t, n\}$. Note that $\mathbf{z} \in \mathbf{Z}$ is a vector of length $T \times N$. This projection step is done by finding the closest feasible point in $\mathbf{Z}$ to the current value $\mathbf{z}^i$, which is formulated as:
\begin{align*}
    \min\nolimits_{\mathbf{z}\in \mathbf{Z}} ||\mathbf{z}^i - \mathbf{z}||_2
\end{align*}
which is a convex optimization problem and therefore can be solved optimally using standard solvers.
This update process continues until convergence, where convergence means that the update does not improve the attacker utility in Eq.~(\ref{dec.0}). Once converged, we obtain a local optimal solution of (\ref{dec.0}--\ref{dec.3}). By running the PGD multiple times with different initial values of the attack variables, we get multiple local optimal solutions. The final solution will be the best with the highest accumulated utility for the attacker among the local optimal ones. The main technical issue is computing the gradients required for PGD, which we discuss next.

The core of PGD is to compute the gradient of the attacker utility at a value $\mathbf{z}$ of the attack variables:
\begin{align*}
     &\frac{dF}{d\mathbf{z}} = \sum\nolimits_{t} \frac{dU^a(\mathbf{x}_{t},\mathbf{z}_{t})}{d\mathbf{z}} \\
    &=\sum\nolimits_{t}\sum\nolimits_{n}\frac{d z_{t,n}}{d \mathbf{z}} U^a_{n}(x_{t,n}) +  z_{t,n}(P^a_{n} - R^a_{n})\frac{d x_{t,n}}{d \mathbf{z}}
\end{align*}
which depends on the two gradient components $\frac{d z_{t,n}}{d \mathbf{z}}$ and $\frac{d x_{t,n}}{d \mathbf{z}}$. The first component, $\frac{d z_{t,n}}{d \mathbf{z}}$, is the gradient of the number of attacks at each target and time step with respect to other targets and steps, $\frac{d z_{t,n}}{d \mathbf{z}} = \big\{\frac{\partial z_{t,n}}{\partial z_{t',n'}} \big\}_{(t',n') \in [T] \times [N]}$, which is straightforwardly determined as follows:
\begin{align*}
    & \frac{\partial z_{t,n}}{\partial z_{t',n'}} = 0 \text{ if } t\neq t' \text{ or } n'\neq n\\
    & \frac{\partial z_{t,n}}{\partial z_{t',n'}} = 1, \text{ otherwise}.
\end{align*}
The second component is $\frac{d x_{t,n}}{d \mathbf{z}} = \big\{\frac{\partial x_{t,n}}{\partial z_{t',n'}}\big\}_{(t',n') \in [T] \times [N]}$, which is the gradient of the defender's strategy at each time step with respect to the number of attacks across all targets and time steps. Note that $\frac{\partial x_{t,n}}{\partial z_{t',n'}}$ is non-zero only when $t' < t$ since the defender's strategy at each step only depends on the historical attacks at previous time steps. By applying the chain rule, it can be decomposed into two parts:
\begin{align*}
    & \frac{\partial x_{t,n}}{\partial z_{t',n'}} =\sum\nolimits_j \frac{\partial x_{t,n}}{\partial \theta_{t,j}} \cdot \frac{\partial\theta_{t,j}}{\partial z_{t',n'}},\forall t' < t\\
    & \frac{\partial x_{t,n}}{\partial z_{t',n'}} = 0,\forall t'\geq t
\end{align*}
where $\theta_{t,j}$ is the $j^{th}$ component of the model parameter vector $\theta_t$ at step $t$. Next, the challenge is that, even though $x_{t,n}$ depends on $\theta_{t}$ and $\theta_{t}$ depends on $z_{t',n'}$, we do not have a closed form of $x_{t,n}$ and $\theta_{t}$ as a function of $\theta_{t}$ and $z_{t',n'}$, respectively. To address this, we take inspiration from hyper-parameter learning~\cite{maclaurin2015gradient}; for utilizing hyper-parameter learning, the attacker has to assume knowledge of the computations steps and model used by defender to solve the problem in Eq.~(\ref{dec.1}) and (\ref{dec.2}). The computational steps can be any differentiable steps that leads to the optimally solving problem in Eq.~(\ref{dec.1}) and (\ref{dec.2}). Here we take the computation by the defender to be a projected gradient descent approach. We show later in our experiments that even when the attacker's assumption about the computation steps is different from the defender's actual model, the attacker still gains a significant benefit for its manipulation.  In the following, we elaborate method to estimate the gradient components $\frac{d \mathbf{x}_t}{d \theta_t}=\{\frac{\partial x_{t,n}}{\partial \theta_{t,j}}\} $ and $ \frac{d\theta_{t}}{d\mathbf{z}_{t'}} = \{\frac{\partial\theta_{t,j}}{\partial z_{t',n'}}\}$ for all $t' < t$.\footnote{In~\cite{thanh2020}, they propose a different approach to compute these gradient components in one-shot SSGs by exploiting intrinsic properties of the defender's learning. Their approach is applicable only when Quantal Response is used. Our approach can be applied for any differentiable behavior models. }
\subsection{Computing Gradient of Defender Strategy }
As mentioned previously, computing the gradient $\frac{d \mathbf{x}_t}{d \theta_t}$ is not straightforward since we do not have a closed-form representation of the defender's strategy $\mathbf{x}_t$ as a function of the behavior model parameter $\theta_t$. Essentially, the defender strategy at time step $t$, $\mathbf{x}_t$, is an optimal solution of:
\begin{align*}
    \max\nolimits_{\mathbf{x}\in\mathbf{X}}\;& U^d(\mathbf{x},\theta_t) 
\end{align*}
The above problem is in general a non-convex optimization problem. Our technique of obtaining $\frac{d \mathbf{x}_t}{d \theta_t}$ is to differentiate through the steps of the defender's PGD approach to solve this problem; this approach is related to the hyper or (sometimes) meta gradient approach in literature~\cite{bengio2000gradient}.

Essentially, the defender starts with an initial strategy $\mathbf{x}^{0,\text{proj}}\in\mathbf{X}$ which is randomly generated. At each iteration of the defender's projected gradient descent $i\geq 1$, given the current defender strategy $\mathbf{x}^{i-1,\text{proj}}$, the  update is: 
\begin{align}\label{eq.1}
    & \mathbf{x}^{i} = \mathbf{x}^{i-1,\text{proj}} + \alpha \frac{\partial U^d(\mathbf{x}^{i-1,\text{proj}},\theta_t)}{\partial \mathbf{x}^{i-1,\text{proj}}}
\end{align}
Then the updated (possibly infeasible) strategy is projected back to the feasible region. We obtain a new feasible strategy $\mathbf{x}^{i,\text{proj}}$ which is an optimal solution of: 
\begin{align}\label{project.1}
    \mathbf{x}^{i,\text{proj}}\in \arg\min\nolimits_{\mathbf{x}\in\mathbf{X}}\; &||\mathbf{x} - \mathbf{x}^{i}||_2
\end{align}
The problem (\ref{project.1}) is a convex optimization problem, which can be easily solved using any convex solver. Clearly, $\mathbf{x}^{i,\text{proj}}$ is a function of $\mathbf{x}^{i}$. We thus have the gradient decomposition:
\begin{align}\label{decomposition}
    & \frac{d\mathbf{x}^{i,\text{proj}}}{d\theta_t} = \frac{d\mathbf{x}^{i,\text{proj}}}{d\mathbf{x}^{i}}\cdot \frac{d\mathbf{x}^{i}}{d\theta_t}
\end{align}
We present Proposition~\ref{prop.1} which shows the computation of the two gradient components on the RHS of (\ref{decomposition}).\footnote{All detailed proofs are in the appendix.}
\begin{proposition}\label{prop.1}
Denote by $G(\mathbf{x}^{i-1,\text{proj}},\theta_t) \!=\!\frac{\partial U^d(\mathbf{x}^{i-1,\text{proj}},\theta_t)}{\partial \mathbf{x}^{i-1,\text{proj}}}$. Denote by $J_{G, \theta_t}$ the sub-matrix of the Jacobian $J_G$ of $G$ that is formed by partial derivatives w.r.t. $\theta_{t,j}$. Similarly, $J_{G, \mathbf{x}^{i-1,\text{proj}}}$ is the sub-matrix of $J_G$ that is formed by partial derivatives w.r.t. $\mathbf{x}^{i-1,\text{proj}}$. Note that then, $J_G = [J_{G, \mathbf{x}^{i-1,\text{proj}}} ~\vert~ J_{G, \theta_t}]$. Then, 
the gradient components in (\ref{decomposition}) are computed as follows:
\begin{align}\label{prop1.1}
     &\frac{d\mathbf{x}^{i}}{d\theta_t} \!=\! \alpha J_{G, \theta_t} \!+\! \left[\alpha J_{G, \mathbf{x}^{i-1,\text{proj}}} \!+\! diag(\vec{1})\right]\cdot \frac{d \mathbf{x}^{i-1,\text{proj}}}{d\theta_t}\\
    &\begin{bmatrix}
       \frac{d\mathbf{x}^{i,\text{proj}}}{d\mathbf{x}^{i}}           \\[0.3em]\label{prop1.2}
       \frac{d \eta}{d\mathbf{x}^{i}}           
\end{bmatrix}
\!=\!-\!\!\begin{bmatrix}
       \nabla_{\mathbf{x}^{i,\text{proj}}}^2 ||\mathbf{x}^i \!-\! \mathbf{x}^{i,\text{proj}}||_2 &\!\!\!\!\!\! \mathbf{A}^T          \\[0.3em]
       diag(\eta)\mathbf{A} &\!\!\!\!\!\! diag(\mathbf{A}\mathbf{x}^{i,\text{proj}} \!-\!b)            
\end{bmatrix}^{-1} \cdot\begin{bmatrix}
       \frac{d\nabla_{\mathbf{x}^{i,\text{proj}}} ||\mathbf{x}^i - \mathbf{x}^{i,\text{proj}}||_2)}{d\mathbf{x}^{i}}           \\[0.3em]
       0            
\end{bmatrix}
\end{align}
\end{proposition}
Essentially, Eq.~(\ref{prop1.1}) is derived by differentiating both sides of Eq.~(\ref{eq.1}). And Eq.~(\ref{prop1.2}) is the result from applying the Implicit Function Theorem~\cite{krantz2012implicit,rudin1986principles} upon the KKT conditions~\cite{boyd2004convex} for the convex problem given by  Eq.~(\ref{project.1}). Here, $\eta$ is the dual variable of $\mathbf{x}^{i,\text{proj}}$. We now present Algorithm~\ref{algorithm.1} which computes $\frac{d\mathbf{x}_t}{d\theta_t}$. Overall, we run $nRound$, each round finds a local optimal strategy solution and its gradient with respect to $\theta_t$. At each round, Algorithm~\ref{algorithm.1} starts by initializing a defender strategy $\mathbf{x}^{0,\text{proj}}$. Then at each iteration $i$, the algorithm updates the defender's strategy as well as its corresponding gradient. This iteration process stops when the update does not increase the defender's utility (i.e., $\delta U \leq 0$). Finally, the optimal defender's strategy and its gradient is determined based on his maximum utility over all rounds (line (\ref{line.11})). 
\begin{algorithm}[t!]
\caption{Compute the gradient $\frac{d \mathbf{x}_t}{d \theta_t}$\label{algorithm.1}}
Initialize $optU = -\infty$;\label{line.1}\\
\For{$round = 1 \to nRound$}{
Initialize $\mathbf{x}^{0,\text{proj}}$; $\delta U = + \infty$; $i = 0$;\label{line.2}\\
\While{$\delta U > 0$}{
    Update $i = i + 1$;\label{line.3}\\
    Compute $\mathbf{x}^i$ and $\mathbf{x}^{i,\text{proj}}$ according to (\ref{eq.1}--\ref{project.1});\label{line.4}\\
    Compute $\frac{d\mathbf{x}^{i,\text{proj}}}{d\theta_t}$ based on (\ref{decomposition}) and Prop.~\ref{prop.1}; \\
    Update $\delta U \!=\! U^d\!(\mathbf{x}^{i,\text{proj}},\theta_t) \!-\! U^d\!(\mathbf{x}^{i-1, \text{proj}},\theta_t)$;\label{line.8}
}
\If{$optU < U^d(\mathbf{x}^{i,\text{proj}},\theta_t)$}{
Update $optU = U^d(\mathbf{x}^{i,\text{proj}},\theta_t)$; $\frac{d \mathbf{x}_t}{d\theta_t} = \frac{d \mathbf{x}^{i,\text{proj}}}{d\theta_t}$;\label{line.11}}
}
\end{algorithm}

\subsection{Compute Gradient of Learning Outcome}
In general, computing the gradient $\frac{d \theta_{t}}{d \mathbf{z}_{t'}}$ is challenging since the learning outcome $\theta_{t}$ depends on the entire attack history before $t$. More specifically, the learning outcome, $\theta_t$, is an optimal solution of the following optimization problem:
\begin{align}\label{mle}
    \min\nolimits_{\theta\in\Theta}\; L(\textbf{X}_{t-1}, \textbf{Z}_{t-1}, \theta)
\end{align}
where $\mathbf{X}_{t-1} = \{\mathbf{x}_1,\dots, \mathbf{x}_{t-1}\}$ and $\mathbf{Z}_{t-1} = \{\mathbf{z}_1,\dots, \mathbf{z}_{t-1}\}$ are the defender's strategies and the attacker's attacks at previous time steps. We consider the set $\Theta$ to be represented by a set of linear constraints $\Theta = \{\theta: \mathbf{\mathbf{C}\cdot \theta \leq \mathbf{D}}\}$. 
This optimization problem is generally known to be non-convex. Similar to the computation of the gradient of the defender strategy, we also apply the a recursive approach to solve this problem and differentiate through the gradient steps. That is, we start with some initial value of $\theta$, denoted by $\theta^{0,\text{proj}}$, which is randomly generated. Then at each iteration $i$, given the current value $\theta^{i-1}$, we update: 
\begin{align}\label{temp.1}
    & \theta^{i} = \theta^{i-1, \text{proj}} - \alpha \frac{d L(\mathbf{X}_{t-1}, \mathbf{Z}_{t-1}, \theta^{i-1,\text{proj}})}{d\theta^{i-1,\text{proj}}}
\end{align}
Then the updated (possibly infeasible) learning outcome $\theta^{i}$ is projected back to the feasible region $\Theta$. We obtain a new feasible learning outcome $\theta^{i,\text{proj}}$ which is the optimal solution of the following minimization problem
\begin{align}\label{model.project}
    \theta^{i,\text{proj}}\in \arg\min\nolimits_{\theta\in\Theta} ||\theta - \theta^{i}||_2
\end{align}
We thus have the following gradient decomposition:
\begin{align}\label{model.grad}
    & \frac{d\theta^{i,\text{proj}}}{d\mathbf{z}_{t'}} = \frac{d\theta^{i,\text{proj}}}{d\theta^{i}}\cdot \frac{d\theta^{i}}{d\mathbf{z}_{t'}}
\end{align}
Observing that the problem (\ref{model.project}) is similar to problem Eq.~(\ref{project.1}), we can thus compute the gradient $\frac{d\theta^{i,\text{proj}}}{d\theta^{i}}$ similarly to $\frac{d\mathbf{x}^{i,\text{proj}}}{d\mathbf{x}^{i}}$. 
On the other hand, the gradient $\frac{d\theta^{i}}{d\mathbf{z}_{t'}}$ is challenging to compute since $\theta^i$ depends on the entire history $(\mathbf{X}_{t-1}, \mathbf{Z}_{t-1})$. We present our Proposition~\ref{prop.2} which shows that this gradient component can be computed recursively according to time steps. 
\begin{proposition}\label{prop.2}
Let $H(\mathbf{X}_{t-1}, \mathbf{Z}_{t-1}, \theta^{i-1,\text{proj}}) = \frac{d L(\mathbf{X}_{t-1}, \mathbf{Z}_{t-1}, \theta^{i-1,\text{proj}})}{d\theta^{i-1,\text{proj}}}$. Let $J_H$ be the Jacobian of $H$. Let $J_{H, x_{t''}}$ be the part of the Jacobian restricted to partial derivatives w.r.t. $x_{t''}$ (and similar for other variables $z_{t'}, \theta^{i-1,\text{proj}}$). The gradient component, $\frac{d\theta^{i}}{d\mathbf{z}_{t'}}$, can be computed recursively, as follows:
\begin{align}\label{temp.2}
    &\frac{d\theta^{i}}{d\mathbf{z}_{t'}} \!=\! \frac{d\theta^{i-1,\text{proj}}}{d\mathbf{z}_{t'}} - \alpha \left[\sum_{t''=t'+1}^{t-1}\!\!J_{H, x_{t''}}\!\cdot\! \frac{ d\mathbf{x}_{t''}}{d \mathbf{z}_{t'}} + J_{H, z_{t'}} + J_{H, \theta^{i-1,\text{proj}}}\!\cdot\! \frac{d \theta^{i-1,\text{proj}}}{d \mathbf{z}_{t'}}\right]\\\label{temp.3}
    & \text{where }\frac{d\mathbf{x}_{t''}}{d\mathbf{z}_{t'}} = \frac{d\mathbf{x}_{t''}}{d\theta_{t''}}\cdot\frac{d\theta_{t''}}{d\mathbf{z}_{t'}}, t' + 1 \leq t'' \leq t-1
\end{align}
\end{proposition}
Therefore, in order to compute all the derivatives $\frac{d\theta_t}{d\mathbf{z}_{t'}}$ for all $t'< t$, we can recursively call a modified version of  Algorithm~\ref{algorithm.1} for $t = 2,\dots, T$ and $t'=1,\dots, t-1$. The inputs of this modified algorithm include: (i) previous defense strategies $\mathbf{X}_{t-1}$; (ii) previous attacks $\mathbf{Z}_{t-1}$; and (iii) the derivatives $\{\frac{d\theta_{t''}}{d\mathbf{z}_{t'}}\}$ and $\{\frac{d\mathbf{x}_{t''}}{d\theta_{t''}}\}$, for all $t'' > t'$ and $t'' < t$. Note that the derivative $\{\frac{d\mathbf{x}_{t''}}{d\theta_{t''}}\}$ is computed based on Algorithm~\ref{algorithm.1}. The full algorithm is stated in the appendix.
\section{Experiments}
Our experiments are conducted on a High Performance Computing (HPC) cluster, with dual E5-2690v4 (28 cores) processors and 128 GB memory. We use Matlab to implement our algorithms. While the attacker assumes the defender uses PGD for his computation, the defender in our experiments actually uses the interior-point method.
We also investigate the impact of the attacker's incorrect knowledge of the defender's learning models on its manipulation outcomes. We analyze the impact of the attacker's manipulative attacks on both players' utility over the entire time horizon. 

To analyze average performance, we generate multiple game with payoffs randomly within the range $[0, 10]$ for the rewards and the range $[-10, 0]$ for the penalties of players at each target, using covariance games from Gambit~\cite{gambit}. This is the commonly-used approach for generating games in security game literature. The covariance value $r$ allows us to govern the correlation between the attacker and defender's payoffs. In particular, when $r = -1$, the games are zero-sum which means the players' utility objectives are completely opposite. On the other hand, when $r = 0$, the players' payoffs are uncorrelated. We use 10 game instances for each value of $r\in\{-1.0, -0.8, -0.6, -0.4, -0.2, 0\}$ (60 games in total). 
In our games, the maximum number of attacks at each time step is limited to $K = 50$. Each of data points is averaged over $60$ games. In the following, we highlight important results. Additional results can be found in the appendix. 
\subsection{Attacker Behavior Models}
In our study, we consider three different behavior models: QR~\cite{yang11}, SUQR~\cite{nguyen2013analyzing}, and SHARP~\cite{kar2015game}, with an increasing order of model complexity (i.e., QR is the simplest model while SHARP is the most complex among the three). These three models have been validated and applied in both human subject experiments and real-world domains such as ferry protection and wildlife security. In addition, these three models are generic, which are suitable for various security settings, including the standard security game setting used in our evaluation. Note that, in addition to these three models, there are other models which were proposed to predict poacher behavior in the wildlife setting. However, these specific models exploit intrinsic properties of the wildlife domain. Therefore, we do not include these models in our experiments. 

\subsection{Evaluation Results}
\begin{figure}[t!]
    \centering
    \subfigure[QR manipulation, $T=4$]{\includegraphics[width = 0.33\textwidth]{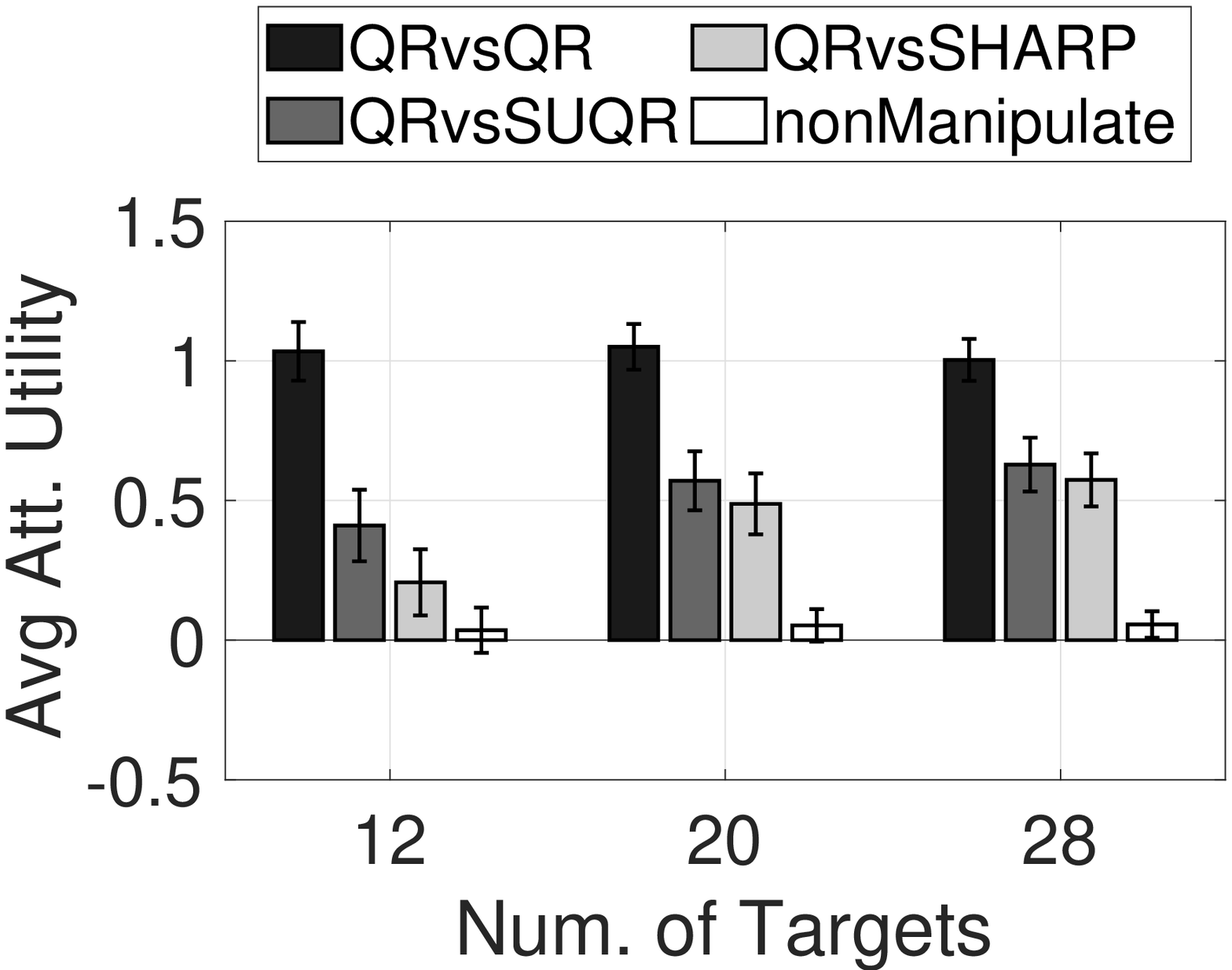}}
    \subfigure[QR manipulation, $T=8$]{\includegraphics[width = 0.33\textwidth]{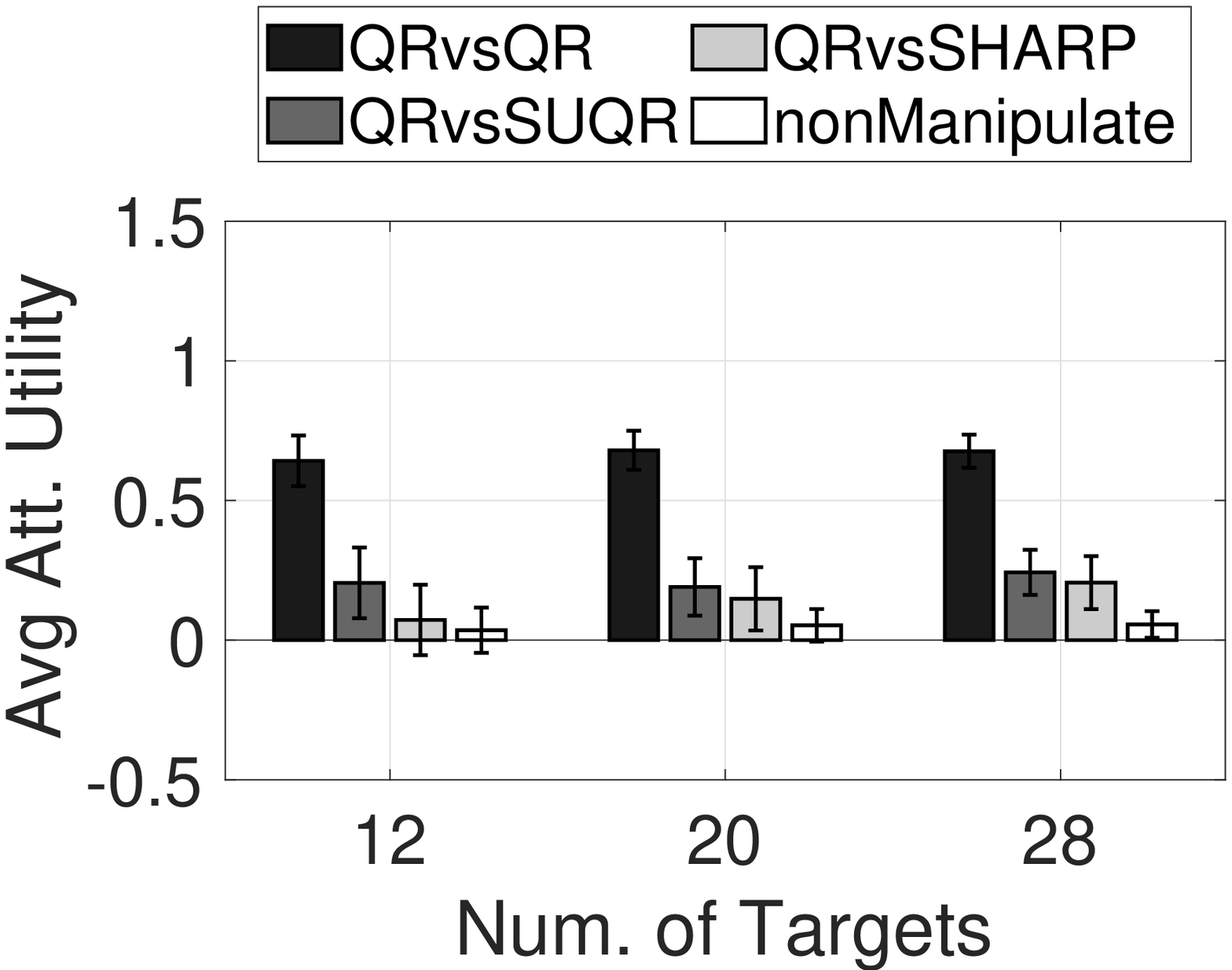}}
    \subfigure[SUQR manipulation, $T=4$]{\includegraphics[width = 0.33\textwidth]{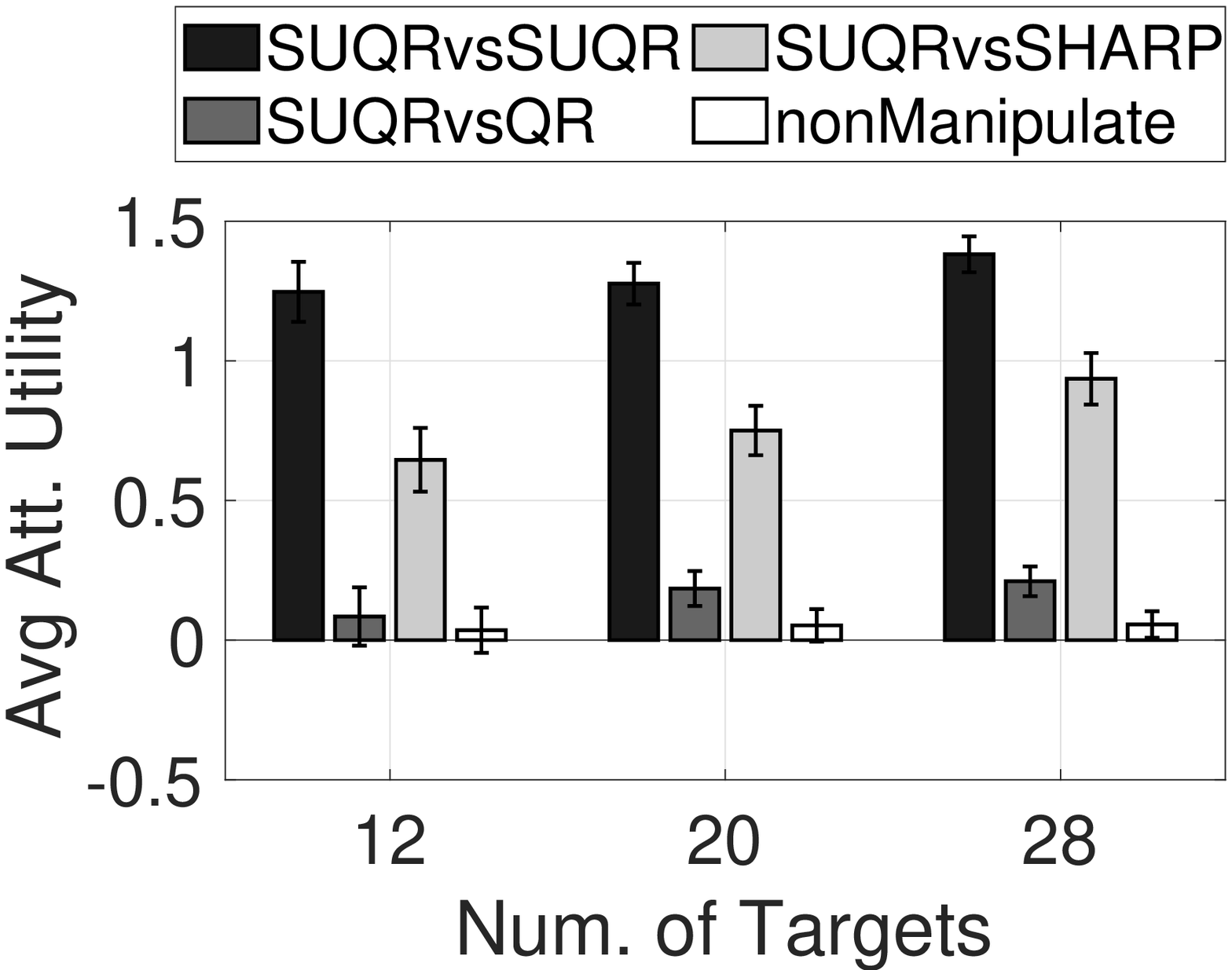}}
    \subfigure[SUQR manipulation, $T=8$]{\includegraphics[width = 0.33\textwidth]{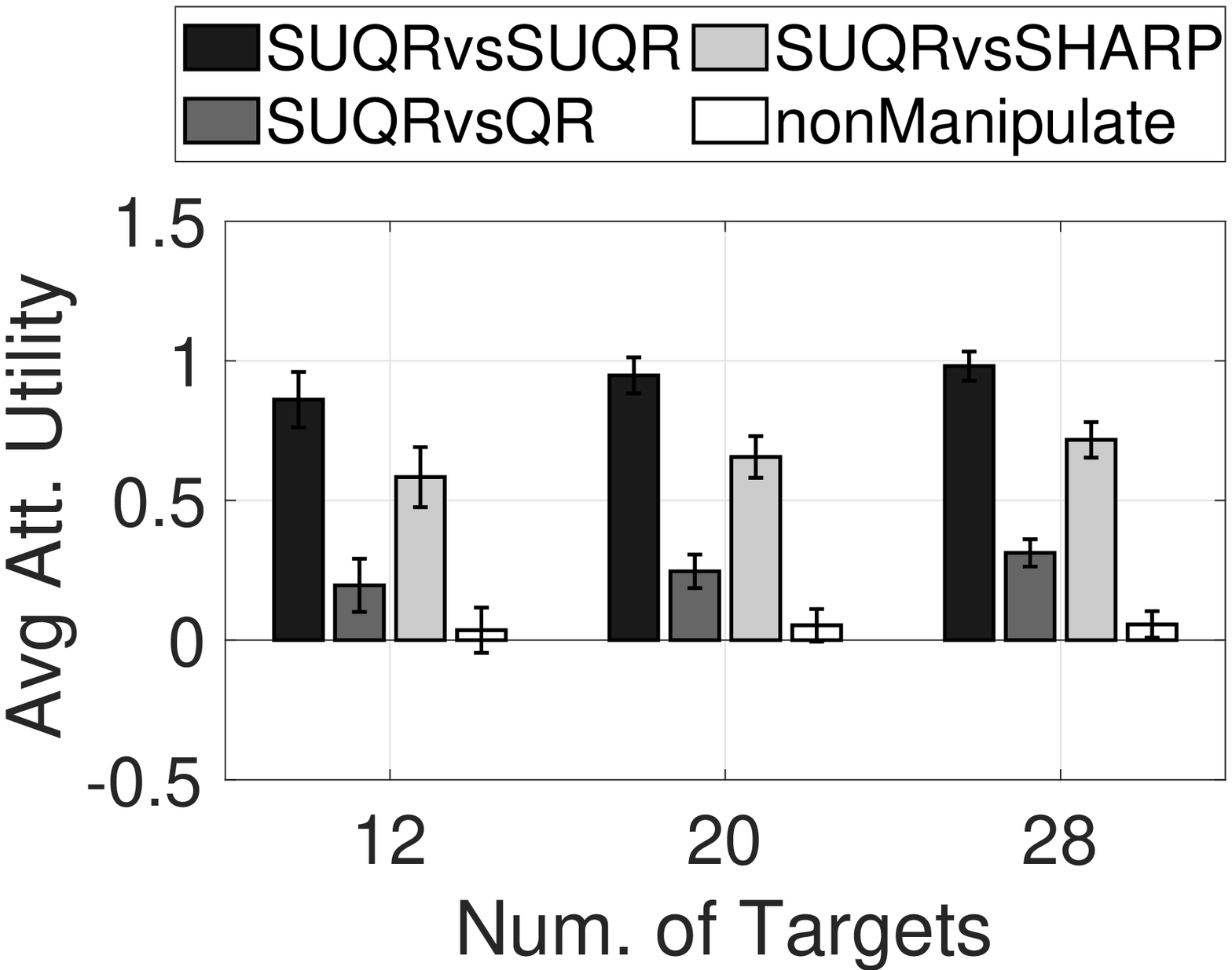}}
    \subfigure[SHARP manipulation, $T\!=\!4$]{\includegraphics[width = 0.33\textwidth]{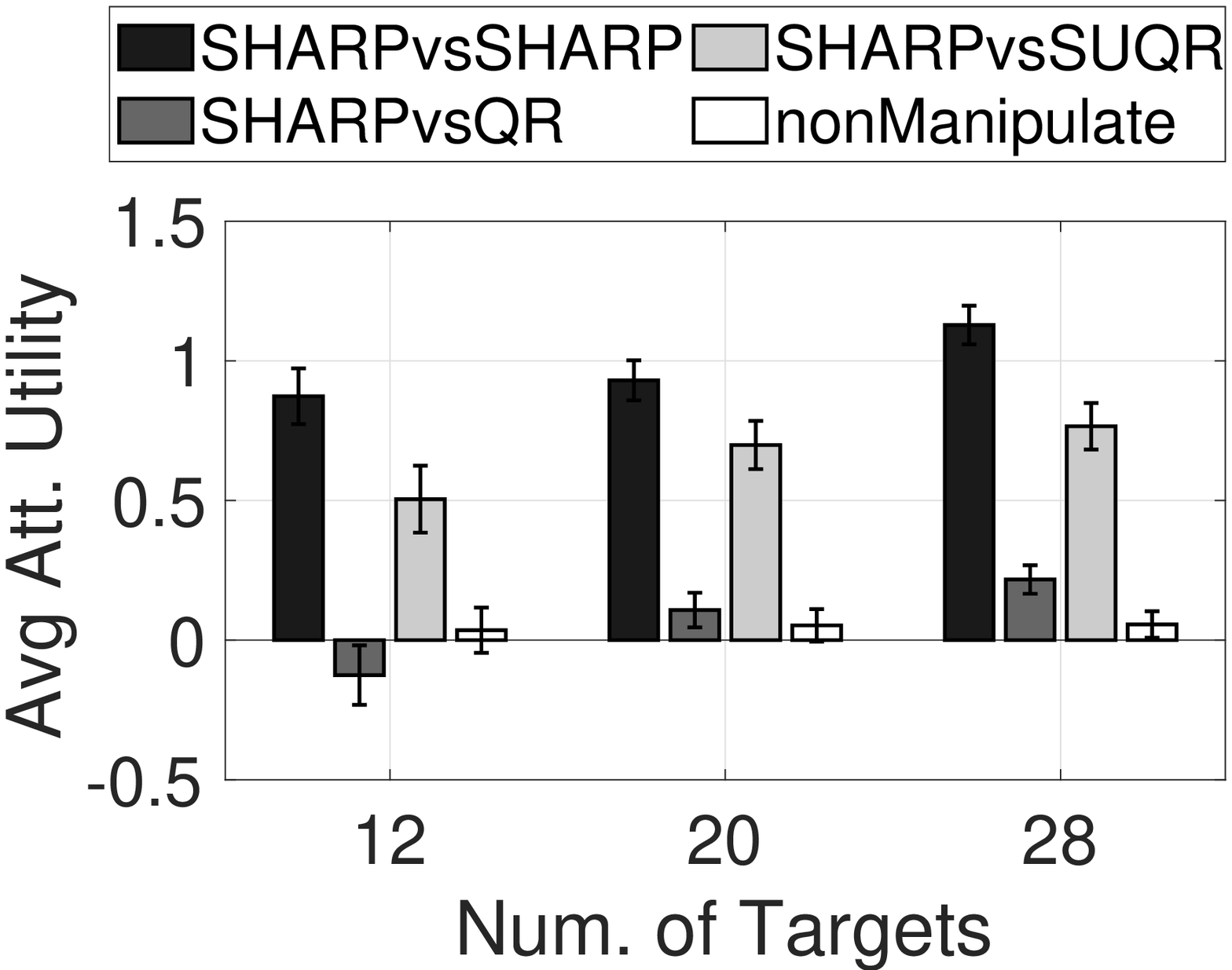}}
    \subfigure[SHARP manipulation, $T\!=\!8$]{\includegraphics[width = 0.33\textwidth]{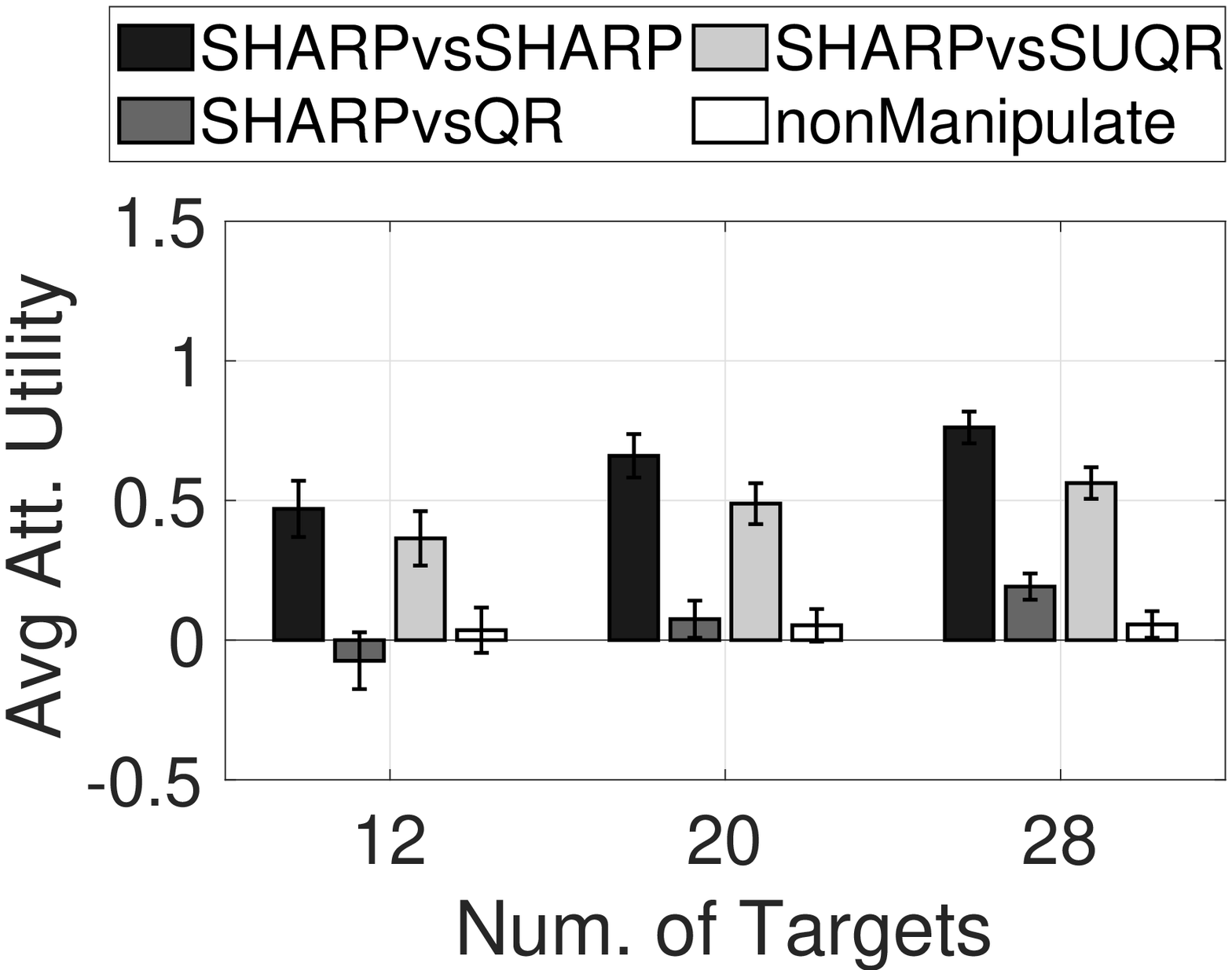}}
    \caption{Attacker Utility Evaluation}
    \label{fig:attU}
\end{figure}

\begin{figure}[t!]
    \centering
    \subfigure[QR manipulation, $T=4$]{\includegraphics[width = 0.33\textwidth]{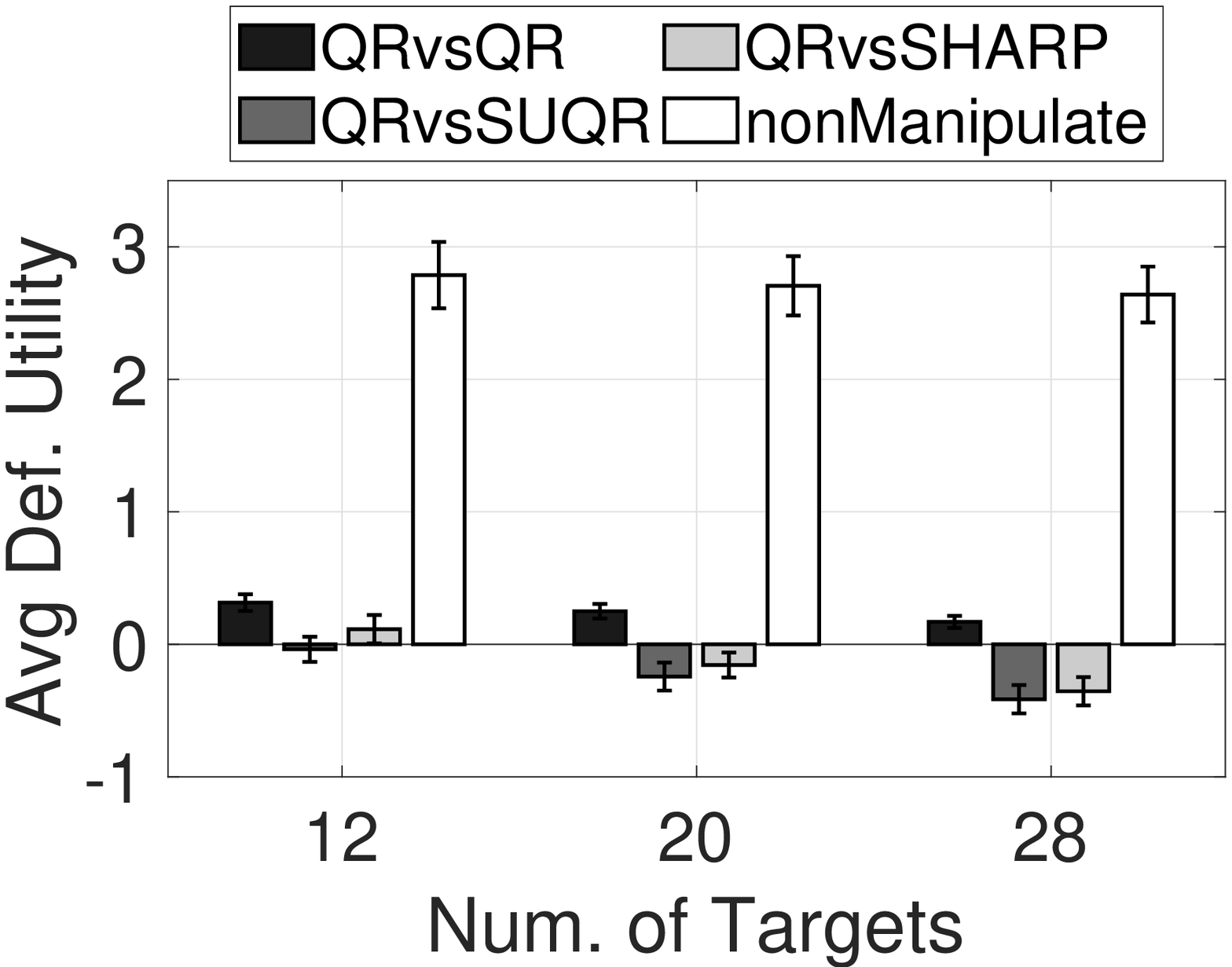}}
    \subfigure[QR manipulation, $T=8$]{\includegraphics[width = 0.33\textwidth]{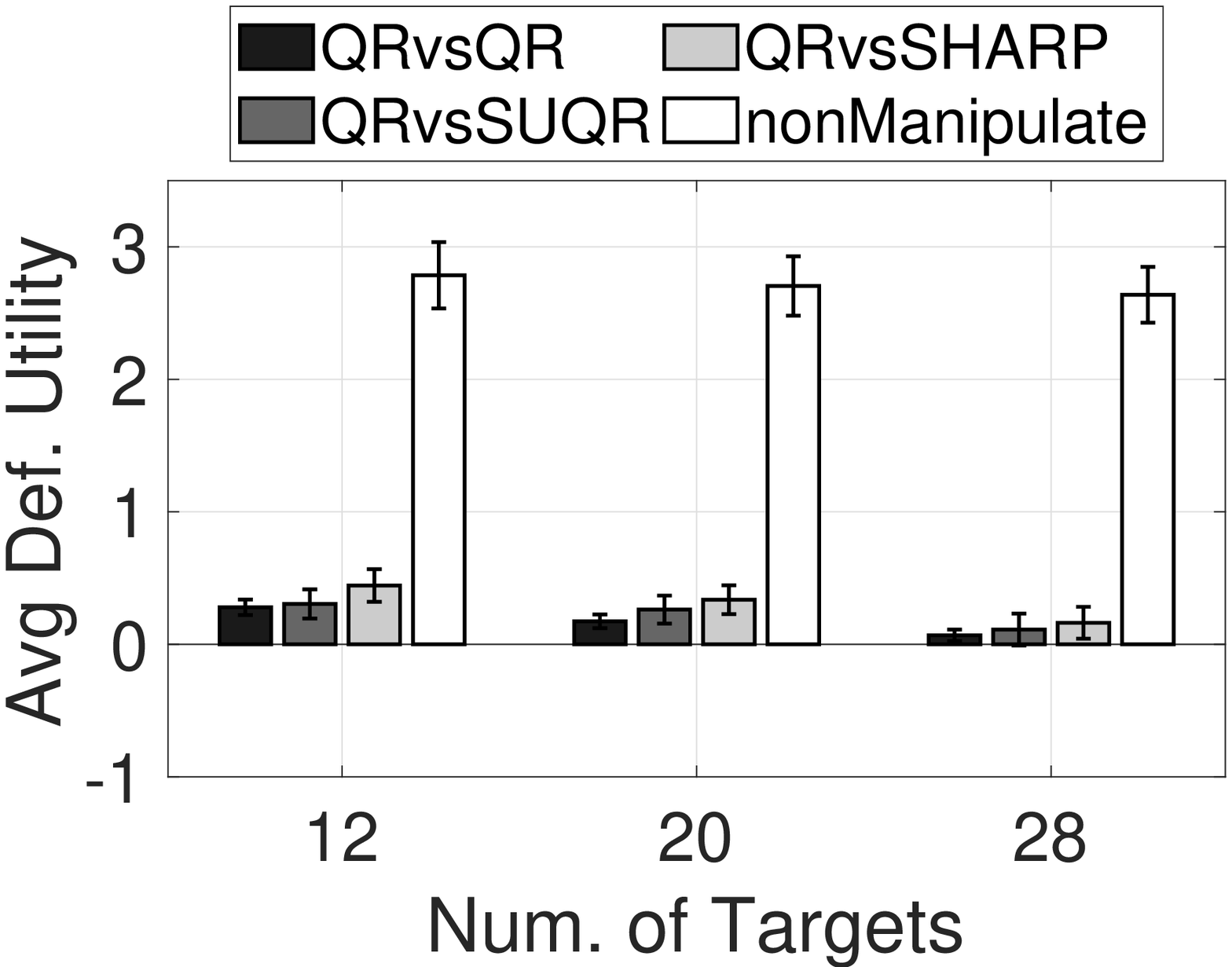}}
    \subfigure[SUQR manipulation, $T=4$]{\includegraphics[width = 0.33\textwidth]{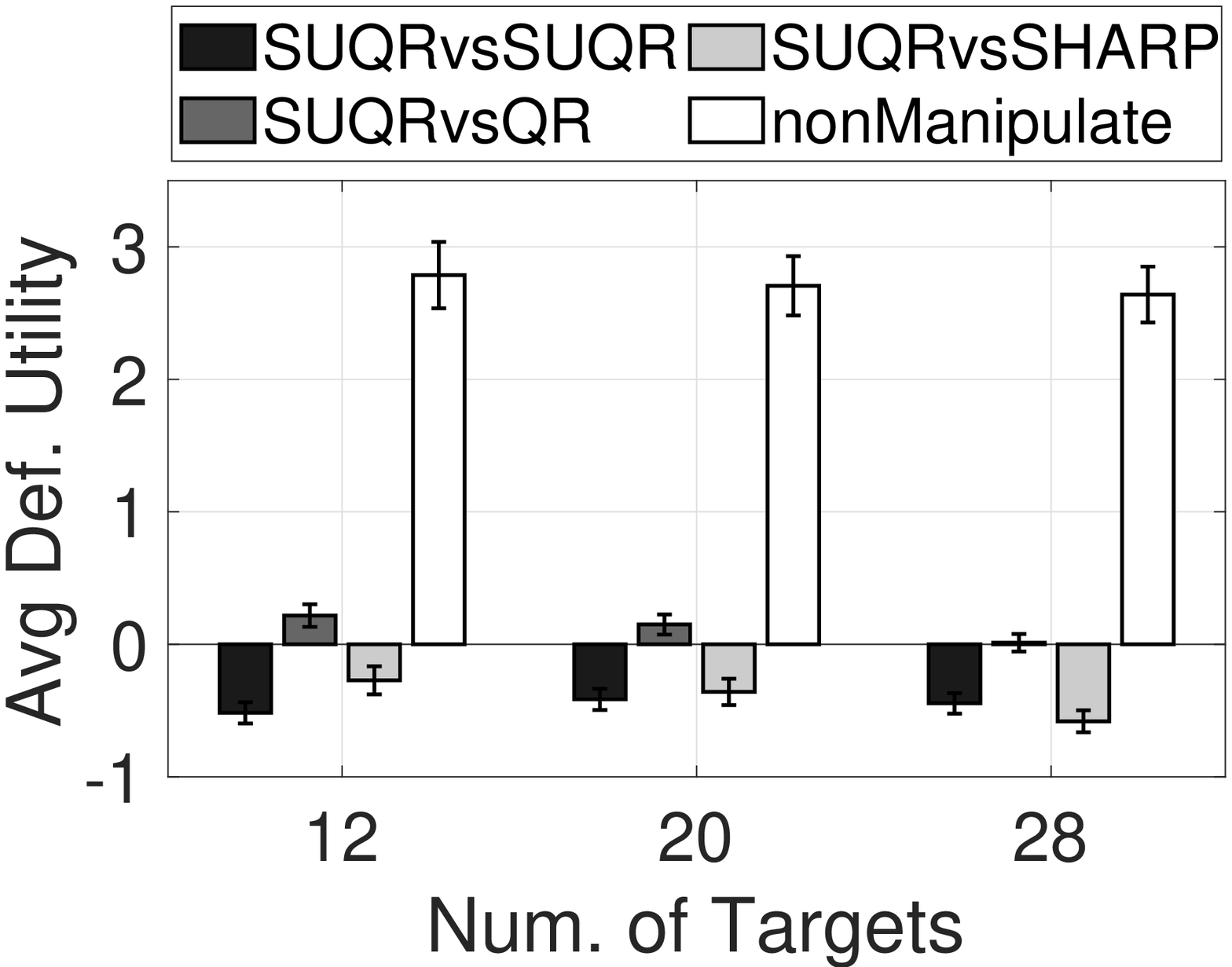}}
    \subfigure[SUQR manipulation, $T=8$]{\includegraphics[width = 0.33\textwidth]{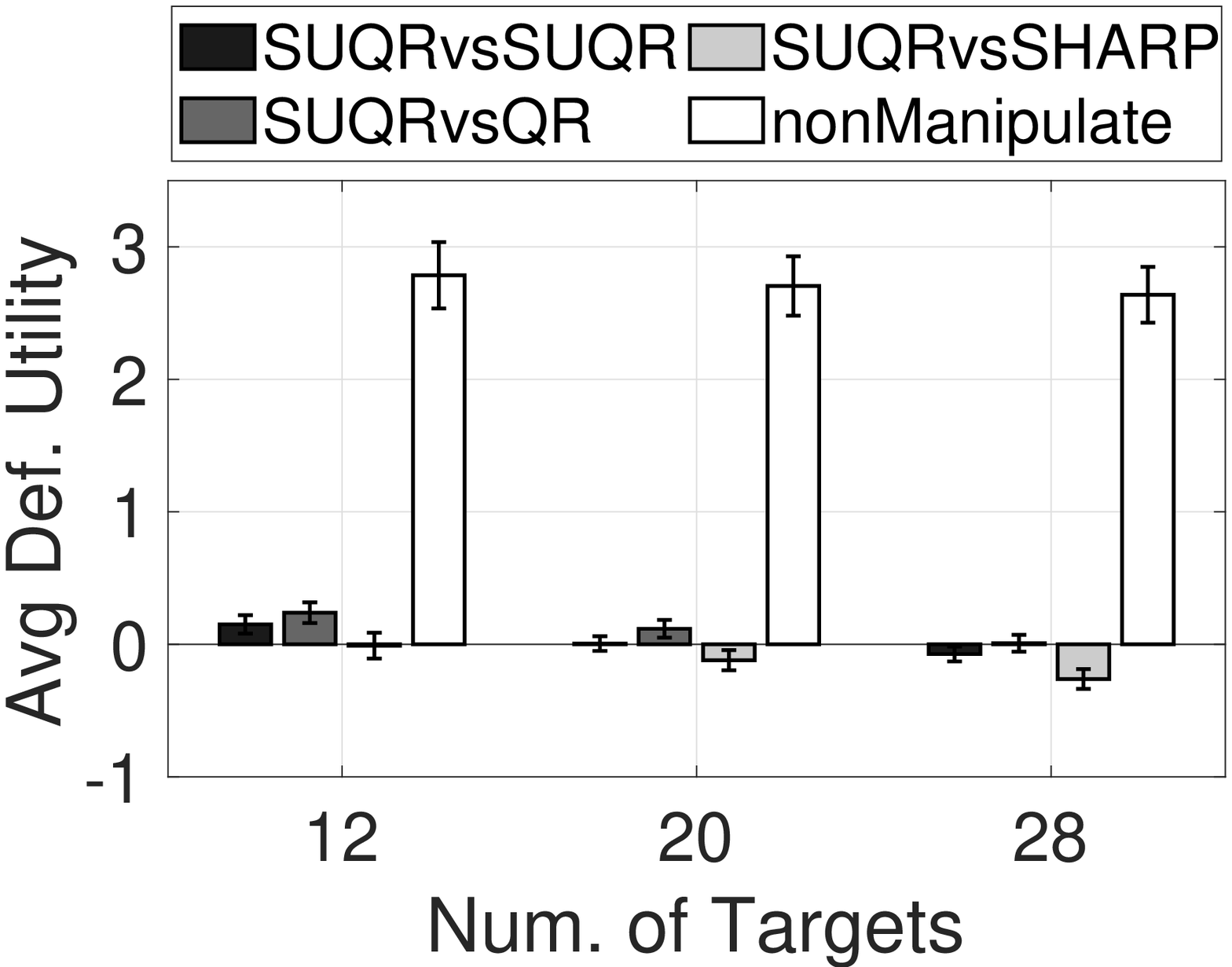}}
    \subfigure[SHARP manipulation, $T\!=\!4$]{\includegraphics[width = 0.33\textwidth]{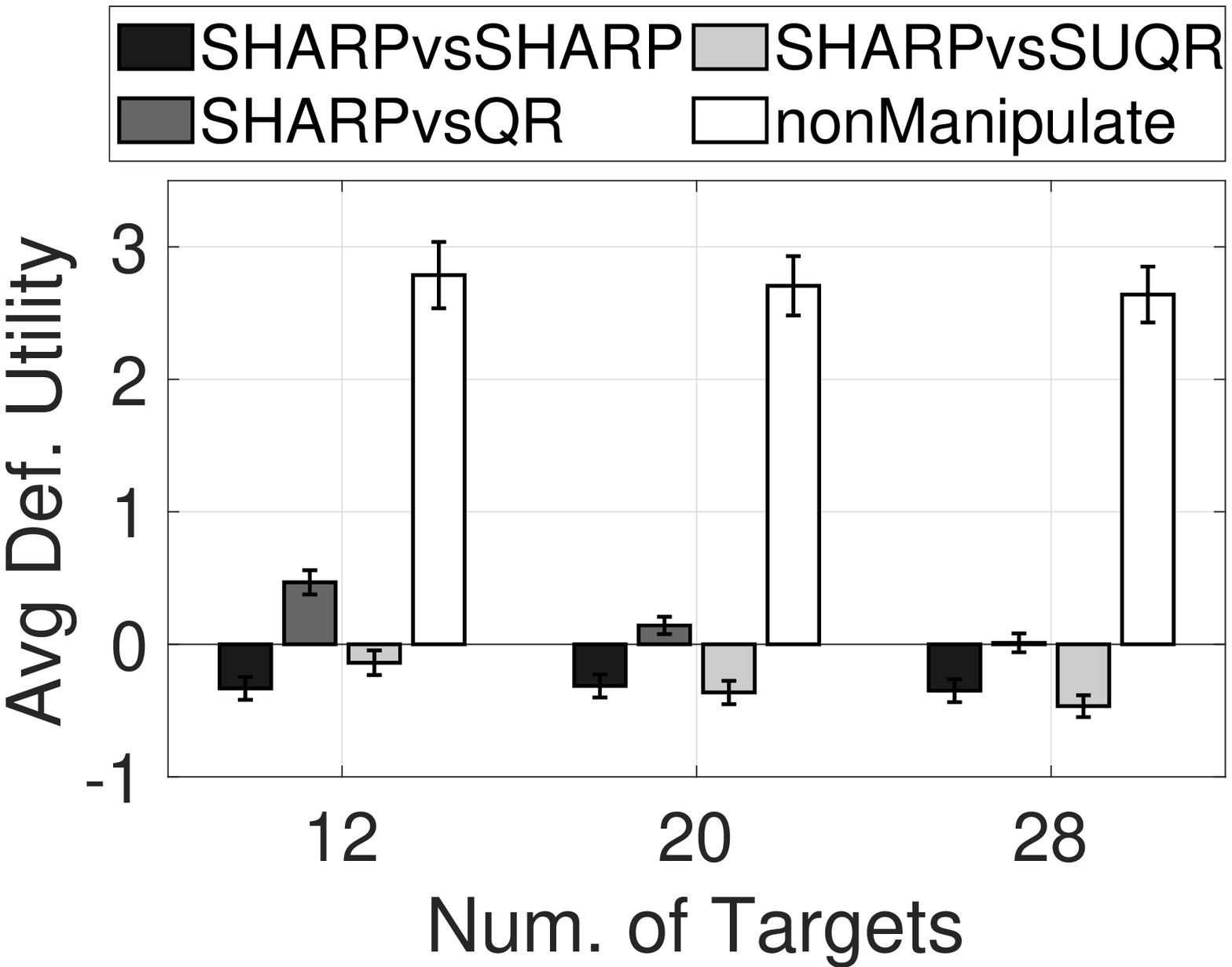}}
    \subfigure[SHARP manipulation, $T\!=\!8$]{\includegraphics[width = 0.33\textwidth]{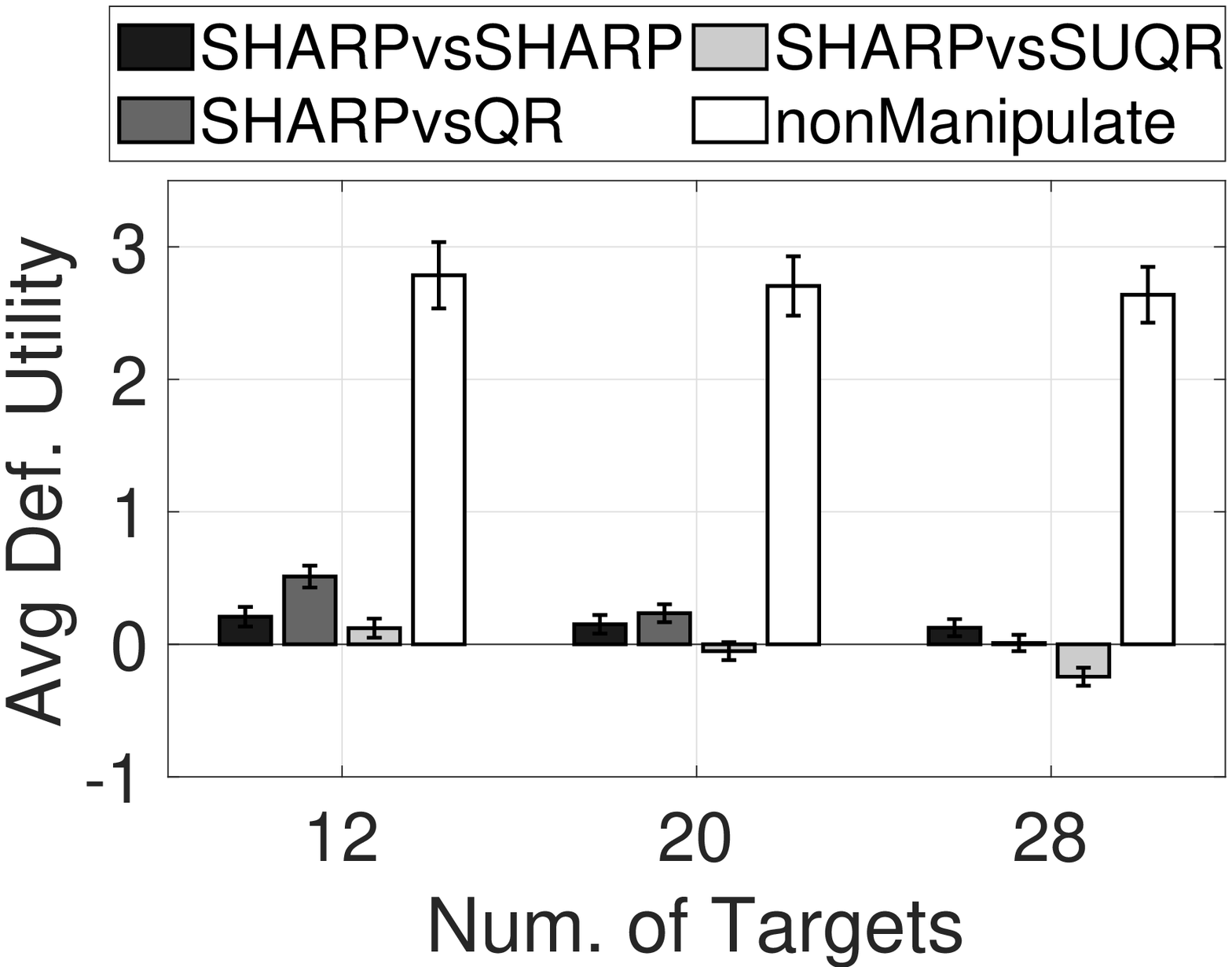}}
    \caption{Defender Utility Evaluation}
    \label{fig:defU}
\end{figure}

As we mentioned previously, the impact of the attacker's manipulative attacks depends on the attacker's knowledge of the defender's learning model. In our experiments, we analyze nine difference manipulation scenarios; each scenario corresponds to a different pair of learning models (a combination of the attacker's assumption of the defender's learning model and the actual model used by the defender among the aforementioned models). In particular, $\mathtt{QRvsQR}$ is the scenario in which (i) the attacker designs its manipulative attacks assuming the defender uses Quantal Response; and (ii) the defender actually uses Quantal Response. and the $\mathtt{QRvsSUQR}$ refers to the scenario in which (i) the attacker assumes the defender uses Quantal Response; and (ii) the defender actually chooses SUQR. The other scenarios are interpreted similarly. Our results on players' utility are shown in Figure~\ref{fig:attU} and Figure~\ref{fig:defU}. In each of these figures, the y-axis is either the attacker or the defender's utility outcomes which are averaged over 60 different games. The x-axis is the number of targets in the games. In addition, $\mathtt{nonManipulate}$ refers to the case when the attacker is not manipulative at all (i.e., both players plays the SSE strategies). For a fair comparison between different number of time steps, we show the averaged utility return per time step. 

Overall, Figure~\ref{fig:attU} shows that the attacker gains a significantly higher utility for manipulating its attacks, especially when the attacker knows the defender's behavior model ($\mathtt{QRvsQR}$, $\mathtt{SUQRvsSUQR}$ and $\mathtt{SHARPvsSHARP}$ in comparison with $\mathtt{nonManipulate}$). When the attacker optimizes its manipulative attacks with respect to a behavior model which is not actually used by the defender, the impact of the attacker's manipulation is much less severe compared to the case of a known behavior model. But in most cases, the attacker's utility for playing manipulatively is still substantially higher than $\mathtt{nonManipulate}$. For example, in Figure~\ref{fig:attU}(a), when the number of targets is $12$, the attacker's average utility is 1.03 in $\mathtt{QRvsQR}$. On the other hand, its utility reduces to 0.41 and 0.20 in $\mathtt{QRvsSUQR}$ and $\mathtt{QRvsSHARP}$, respectively, which is still significantly higher than $\mathtt{nonManipulate}$ with the attacker utility of 0.03. 

There is a notable case in which the attacker suffers loss when it assumes the SHARP model (the most complex one among the three model) while the defender actually uses QR (the simplest model) (Figure~\ref{fig:attU}(e)(f) with 12-target games and $\mathtt{SHARPvsQR}$ versus $\mathtt{nonManipulate}$). Indeed, an overparameterized complex model generally tends to be more sensitive towards noise and has been shown to a poor choice for a surrogate model in transferability of attacks in machine learning~\cite{demontis2019adversarial}. Thus, when the learning model is unknown, it would be beneficial for the attacker to use a simple behavior model.

Next, we examine the long-term benefit of the attacker's deception by comparing its utility between the 4-step games and the 8-step games (Figures~\ref{fig:attU}(a)(c)(e) versus Figures~\ref{fig:attU}(b)(d)(f)). For a fair comparison, we show results of the average utility return per step of the attacker. In these figures, we observe that when the number of time steps increases (8 versus 4 steps), the average utility return per step for the attacker is reduced. This result indicates that the attacker's deception has less impact on its benefit in a longer-term. This result is reasonable since the attacker has to trade off between the deception benefit (for misleading the defender's learning and patrolling) and the immediate utility loss for playing boundedly rational at every time step. And, in a long run, manipulative attacks at later time steps would have less impact on the defender's learning outcome.

Regarding the defender's utility (Figure~\ref{fig:defU}), regardless of whether the attacker knows the defender's choice of a learning behavior model or not, the defender suffers a significant loss in utility in the presence of the attacker's manipulation (compared to the $\mathtt{nonManipulate}$ case). This result clearly shows that when the defender relies on attacker behavior learning, the quality of his strategy outcome is extremely vulnerable to the manipulative attacks of a clever attacker. 

\begin{figure}[t!]
    \centering
    \subfigure[4 steps]{\includegraphics[width = 0.33\textwidth]{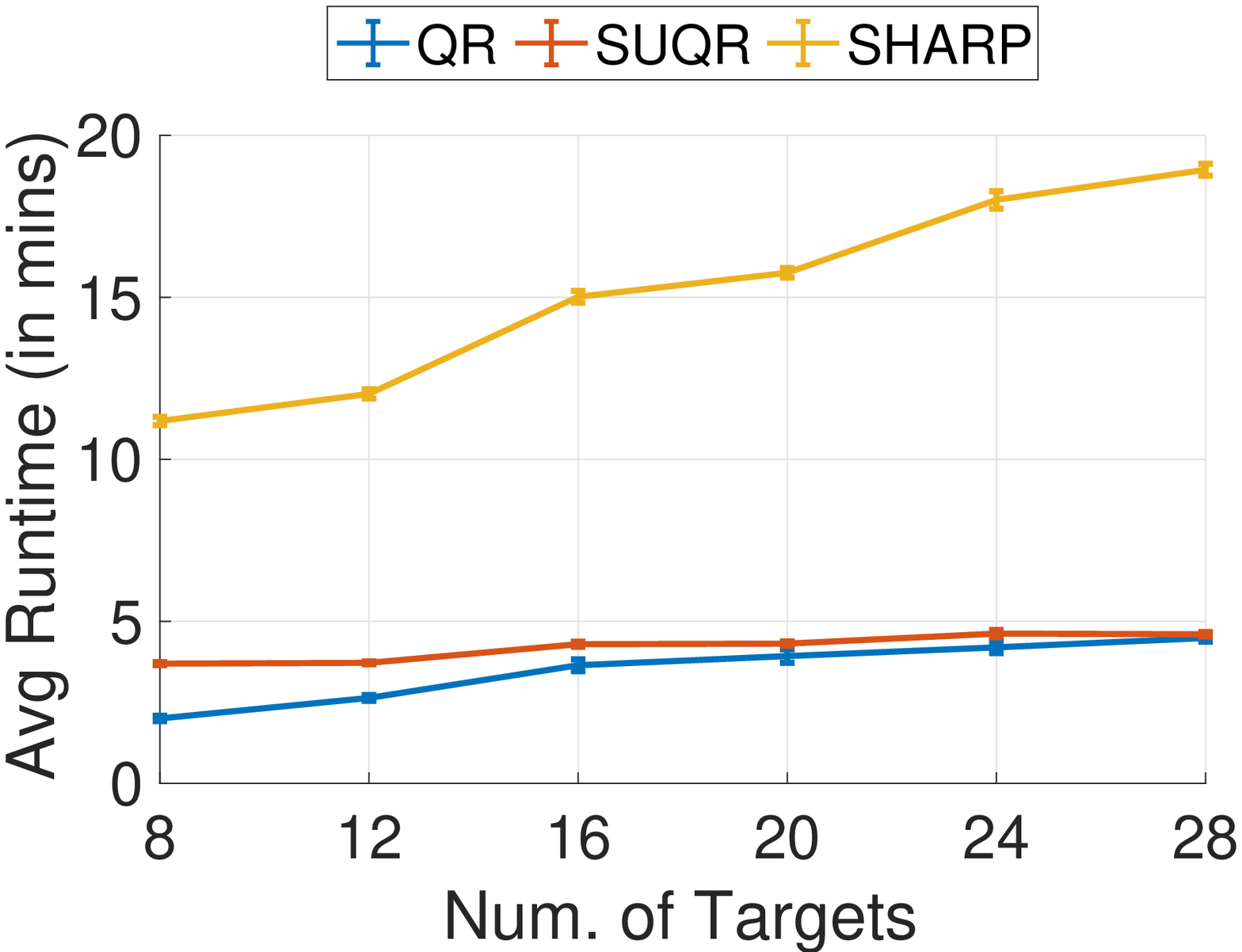}}
    \subfigure[8 steps]{\includegraphics[width = 0.33\textwidth]{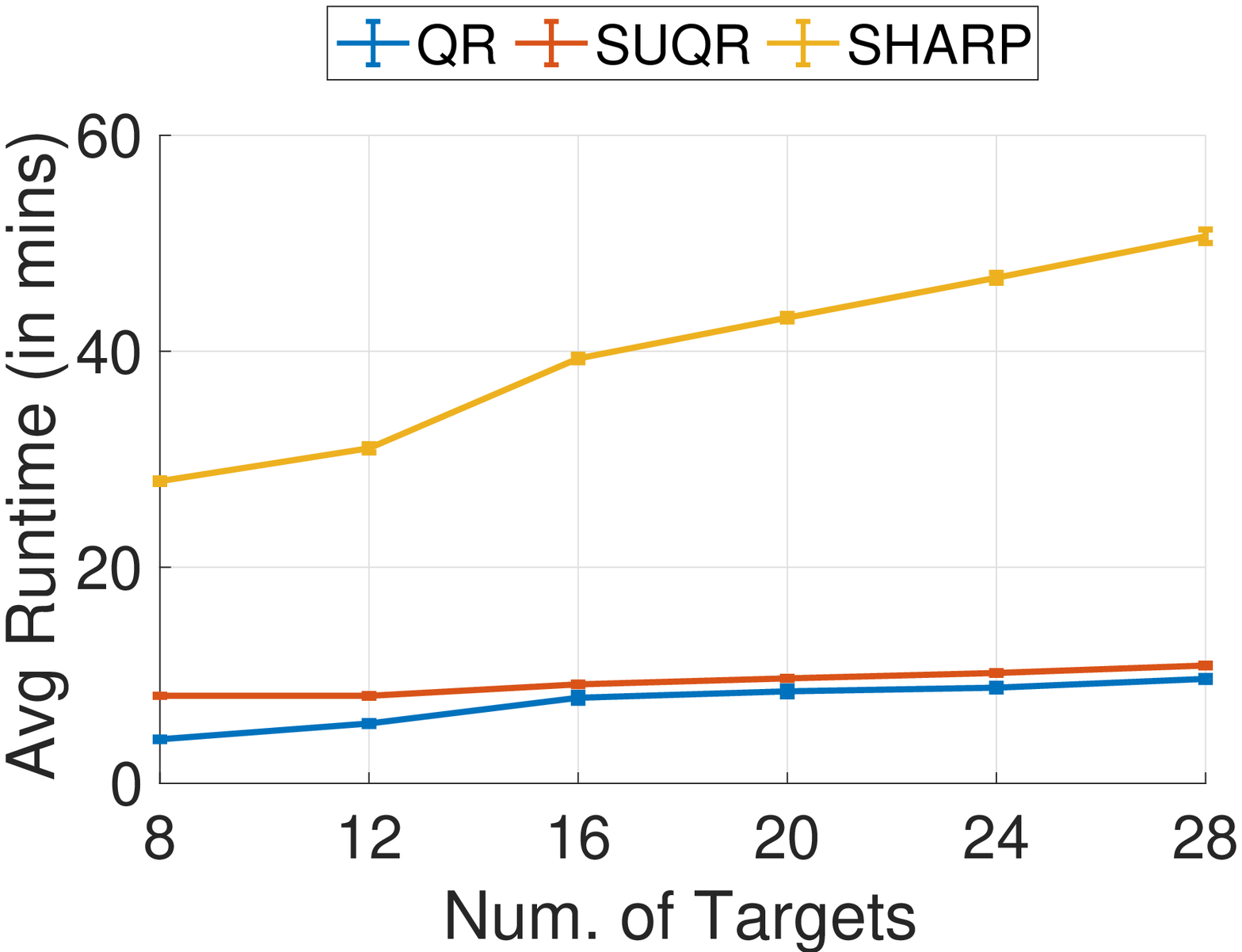}}
    \caption{Runtime Performance}
    \label{fig:runtime}
\end{figure}

Finally, we examine the runtime performance of our proposed algorithm. The result is shown in Figure~\ref{fig:runtime} in which the y-axis is the average runtime in minutes. Each of our data points is computed based on aggregated 5 rounds of the PGD process; each round consists of approximately 30 iterations of gradient descent update until reaching a local optimal solution. Note that each iteration involves multiple optimization components (i.e., training the behavior model and computing an optimal strategy at each time step). Despite the complex computation, Figure~\ref{fig:runtime} shows that the runtime increases linearly in the number of targets, suggesting our method can be scaled up for large games.
\section{Summary}
In this work, we investigate the problem of sequential manipulative attacks in multi-step security games. We formulate new gradient based algorithms to compute an optimal attack plan for the attacker, tackling the computational challenge due to multiple inter-connected optimization components across the entire time horizon. Our experiments in various game settings show that when relying on historical data for learning, the defender's patrol strategies become extremely vulnerable to the attacker's manipulation of attack data, regardless of whether the attacker knows the defender's learning model or not. Our future research goal is to design effective learning-planning solutions for the defender in security games that are resilient to such attacker manipulation.  
\paragraph{Acknowledgement}
This work was supported by ARO grant W911NF-20-1-0344 from the US Army Research Office.

\bibliographystyle{unsrt} 
\bibliography{submission}
\clearpage
\section*{Appendix}
\subsection*{Proof of Proposition~\ref{prop.1}}
\begin{proof}
We denote by $G(\mathbf{x}^{i-1,\text{proj}},\theta_t) = \frac{\partial U^d(\mathbf{x}^{i-1,\text{proj}},\theta_t)}{\partial \mathbf{x}^{i-1,\text{proj}}}$, which is a function of $(\mathbf{x}^{i-1,\text{proj}},\theta_t)$. Recall the notation $J_G = [J_{G, \mathbf{x}^{i-1,\text{proj}}} ~\vert~ J_{G, \theta_t}]$. By taking the derivative on both side of Eq.~(\ref{eq.1}) with respect to $\theta_t$, we obtain:
\begin{align*}
    \frac{d\mathbf{x}^{i}}{d\theta_t} &= \frac{d\mathbf{x}^{i-1,\text{proj}}}{d\theta_t} + \alpha \frac{dG(\mathbf{x}^{i-1,\text{proj}},\theta_t)}{d\theta_t} \\
    & = \frac{d\mathbf{x}^{i-1,\text{proj}}}{d\theta_t} + \alpha \frac{dG(\mathbf{x}^{i-1,\text{proj}},\theta_t)}{d(\mathbf{x}^{i-1,\text{proj}},\theta_t)} \frac{d(\mathbf{x}^{i-1,\text{proj}},\theta_t)}{d\theta_t}) \\
    &= \frac{d\mathbf{x}^{i-1,\text{proj}}}{d\theta_t} + \alpha \left[J_{G,\theta_t} + J_{G, \mathbf{x}^{i-1,\text{proj}}}\cdot\frac{d \mathbf{x}^{i-1,\text{proj}}}{d\theta_t}\right]\\
    \implies \frac{d\mathbf{x}^{i}}{d\theta_t} &= \alpha J_{G, \theta_t} + \left[\alpha J_{G, \mathbf{x}^{i-1,\text{proj}}} + diag(\vec{1})\right]\cdot \frac{d \mathbf{x}^{i-1,\text{proj}}}{d\theta_t}
\end{align*}
which show that we can compute the gradient $\frac{d\mathbf{x}^{i}}{d\theta_t}$ recursively according to the gradient ascent step.

Next, we will describe the computation of $\frac{d\mathbf{x}^{i,\text{proj}}}{d\mathbf{x}^{i}}$, which is the gradient of the projected strategy $\mathbf{x}^{i,\text{proj}}$ with respect to the gradient-based updated strategy $\mathbf{x}^{i}$. Note that the projection problem (\ref{project.1}) is a convex optimization problem:
\begin{align}\label{project.abstract.1}
\min_{\mathbf{x}}\;& f(\mathbf{x},\mathbf{x}^{i'})\\\label{project.abstract.2}
\text{s.t. } & A\mathbf{x} \leq b
\end{align}
We can thus apply the Implicit Function Theorem~\cite{krantz2012implicit,rudin1986principles} upon the KKT conditions~\cite{boyd2004convex} of this problem, to obtain the gradient $\frac{d\mathbf{x}^{i,\text{proj}}}{d\mathbf{x}^{i}}$, formulated as follows: 
\begin{align}\nonumber
 &\begin{bmatrix}
       \nabla_{\mathbf{x}^{i,\text{proj}}}^2 f(\mathbf{x}^{i,\text{proj}},\mathbf{x}^{i}) & \mathbf{A}^T           \\[0.3em]
       diag(\eta)\mathbf{A} & diag(\mathbf{A}\mathbf{x}^{i,\text{proj}} -b)            
\end{bmatrix} 
\begin{bmatrix}
       \frac{d\mathbf{x}^{i,\text{proj}}}{d\mathbf{x}^{i}}           \\[0.3em]
       \frac{d \eta}{d\mathbf{x}^{i}}           
\end{bmatrix}\\
&=-\begin{bmatrix}
       \frac{d\nabla_{\mathbf{x}^{i,\text{proj}}} f(\mathbf{x}^{i,\text{proj}},\mathbf{x}^{i})}{d\mathbf{x}^{i}}           \\[0.3em]
       0            
\end{bmatrix}   \\\nonumber
&\text{which implies: }\\\nonumber &\begin{bmatrix}
       \frac{d\mathbf{x}^{i,\text{proj}}}{d\mathbf{x}^{i}}           \\[0.3em]
       \frac{d \eta}{d\mathbf{x}^{i}}           
\end{bmatrix}=-\begin{bmatrix}
       \nabla_{\mathbf{x}^{i,\text{proj}}}^2 f(\mathbf{x}^{i,\text{proj}},\mathbf{x}^{i}) & \mathbf{A}^T           \\[0.3em]
       diag(\eta)\mathbf{A} & diag(\mathbf{A}\mathbf{x}^{i,\text{proj}} -b)            
\end{bmatrix}^{-1}\\
&\qquad\qquad\quad\cdot\begin{bmatrix}
       \frac{d\nabla_{\mathbf{x}^{i,\text{proj}}} f(\mathbf{x}^{i,\text{proj}},\mathbf{x}^{i})}{d\mathbf{x}^{i}}           \\[0.3em]
       0            
\end{bmatrix}
\end{align}
where $\eta$ is the dual variable with respect to $\mathbf{x}^{i,\text{proj}}$.
\end{proof}
\subsection{Proof of Proposition~\ref{prop.2}}
Let $H(\mathbf{X}_{t-1}, \mathbf{Z}_{t-1}, \theta^{i-1,\text{proj}}) = \frac{d L(\mathbf{X}_{t-1}, \mathbf{Z}_{t-1}, \theta^{i-1,\text{proj}})}{d\theta^{i-1,\text{proj}}}$. By taking the derivatives on both sides of (\ref{temp.1}), we obtain:
\begin{align*}
    & \frac{d\theta^{i}}{d\mathbf{z}_{t'}} = \frac{d\theta^{i-1, \text{proj}}}{d\mathbf{z}_{t'}} - \alpha \frac{dH}{d\mathbf{z}_{t'}}
\end{align*}
We observe that $H$ is a function of $(\mathbf{X}_{t-1}, \mathbf{Z}_{t-1}, \theta^{i-1,\text{proj}})$ in which $\mathbf{x}_{t''}\in \mathbf{X}_{t-1}$ with $t'' \leq t-1$ is a function of $\mathbf{z}_{t'}$ for all $t'' > t'$. In addition, $\theta^{i-1,\text{proj}}$ is also a function of $\mathbf{z}_{t'}$. Therefore, by applying the chain rule, we obtain:  
\begin{align*}
    & \frac{dH}{d\mathbf{z}_{t'}} = \frac{dH}{d(\mathbf{X}_{t-1}, \mathbf{Z}_{t-1}, \theta^{i-1,\text{proj}})} \frac{d(\mathbf{X}_{t-1}, \mathbf{Z}_{t-1}, \theta^{i-1,\text{proj}})}{d\mathbf{z}_{t'}}\\
    & = \sum_{t''=1}^{t-1}\!\!J_{H, x_{t''}}\!\cdot\! \frac{ d\mathbf{x}_{t''}}{d \mathbf{z}_{t'}}\!+\!J_{H, z_{t''}}\!\cdot\! \frac{ d\mathbf{z}_{t''}}{d \mathbf{z}_{t'}} + J_{H, \theta^{i-1,\text{proj}}}\!\cdot\! \frac{d \theta^{i-1,\text{proj}}}{d \mathbf{z}_{t'}}\! 
\end{align*}
Note that $\frac{d \mathbf{z}_{t''}}{d \mathbf{z}_{t'}} = 0 $ if $\mathbf{z}_{t''} \neq \mathbf{z}_{t'}$ and is the identity matrix $I$ otherwise. Also, $\frac{d \mathbf{x}_{t''}}{d \mathbf{z}_{t'}} = 0 $ for all $t'' \leq t'$. Thus, the above is same as
\begin{align*}
    &= \sum_{t''=t'+1}^{t-1}\!\!J_{H, x_{t''}}\!\cdot\! \frac{ d\mathbf{x}_{t''}}{d \mathbf{z}_{t'}} + J_{H, z_{t'}} + J_{H, \theta^{i-1,\text{proj}}}\!\cdot\! \frac{d \theta^{i-1,\text{proj}}}{d \mathbf{z}_{t'}}\!
\end{align*}
Note that when $t' = t-1$, we have:
\begin{align*}
    &\frac{dH}{d\mathbf{z}_{t-1}} =  J_{H, z_{t'}} + J_{H, \theta^{i-1,\text{proj}}}\!\cdot\! \frac{d \theta^{i-1,\text{proj}}}{d \mathbf{z}_{t'}}
\end{align*}
\subsection{Algorithm to Compute Gradient $\frac{d\theta_t}{d\mathbf{z}_{t'}}$}

In Algorithm~\ref{alogirhtm.2}, we recursively compute all the gradients $\frac{d\theta_t}{d\mathbf{z}_{t'}}$ for $t = 2,\dots, T$ and $t'=1,\dots, t-1$. 
\begin{algorithm}
\caption{\label{alogirhtm.2}Compute the gradients $\{\frac{d\theta_t}{d\mathbf{z}_{t'}}$  ($t'< t$)\}}
\For{$t = 2\to N$}{
\For{$t' = 1 \to t - 1$}{
Compute the derivatives $\frac{d\mathbf{x}_{t''}}{d\mathbf{z}_{t'}} = \frac{d\mathbf{x}_{t''}}{d\theta_{t''}}\cdot\frac{d\theta_{t''}}{d\mathbf{z}_{t'}}$ for all $ t' + 1 \leq t'' \leq t-1$ where the derivative $\frac{d\mathbf{x}_{t''}}{d\theta_{t''}}$ is computed by Algorithm~\ref{algorithm.1};\\
Initialize $optL = +\infty$;\\
\For{$round = 1 \to nRound$}{
Initialize $\mathbf{\theta}^{0,\text{proj}}$; $\delta L = + \infty$; $i = 0$;\\
\While{$\delta L > 0$}{
    Update $i = i + 1$;\\
    Compute $\mathbf{\theta}^i, \mathbf{\theta}^{i,\text{proj}}$ based on (\ref{temp.1}--\ref{model.project});\\
    Compute $\frac{d\theta^{i,\text{proj}}}{d\mathbf{z}_{t'}} \!=\! \frac{d\theta^{i,\text{proj}}}{d\theta^{i}}\!\!\cdot\! \frac{d\theta^{i}}{d\mathbf{z}_{t'}}$ where $\frac{d\theta^{i,\text{proj}}}{d\theta^{i}}$ is similar to $\frac{d\mathbf{x}^{i,\text{proj}}}{d\mathbf{x}^{i}}$ and $\frac{d\theta^{i}}{d\mathbf{z}_{t'}}$ depends on $(\!\frac{d\mathbf{x}_{t''}}{d\mathbf{z}_{t'}}, \frac{d\theta^{i\!-\!1,\text{proj}}}{d\mathbf{z}_{t'}}\!)$ (Prop.~\ref{prop.2});\\
    Update $\delta L \!=\! L(\textbf{X}_{t-1}, \textbf{Z}_{t-1}, \theta^{i,\text{proj}}) \!-\! L(\textbf{X}_{t-1}, \textbf{Z}_{t-1}, \theta^{i-1,\text{proj}})$;
}
\If{$optL < L(\textbf{X}_{t-1}, \textbf{Z}_{t-1}, \theta^{i,\text{proj}})$}{
Update $optL \!=\! L(\textbf{X}_{t-1}, \textbf{Z}_{t-1}, \theta^{i,\text{proj}})$; $\frac{d \mathbf{\theta}_t}{d\mathbf{z}_{t'}} = \frac{d \mathbf{\theta}^{i,\text{proj}}}{d\mathbf{z}_{t'}}$;}
}
}
}
\end{algorithm}

\subsection{Additional Experiment Results}
\begin{figure}[t!]
    \centering
    \subfigure[QR manipulation, $T=4$]{\includegraphics[width = 0.33\textwidth]{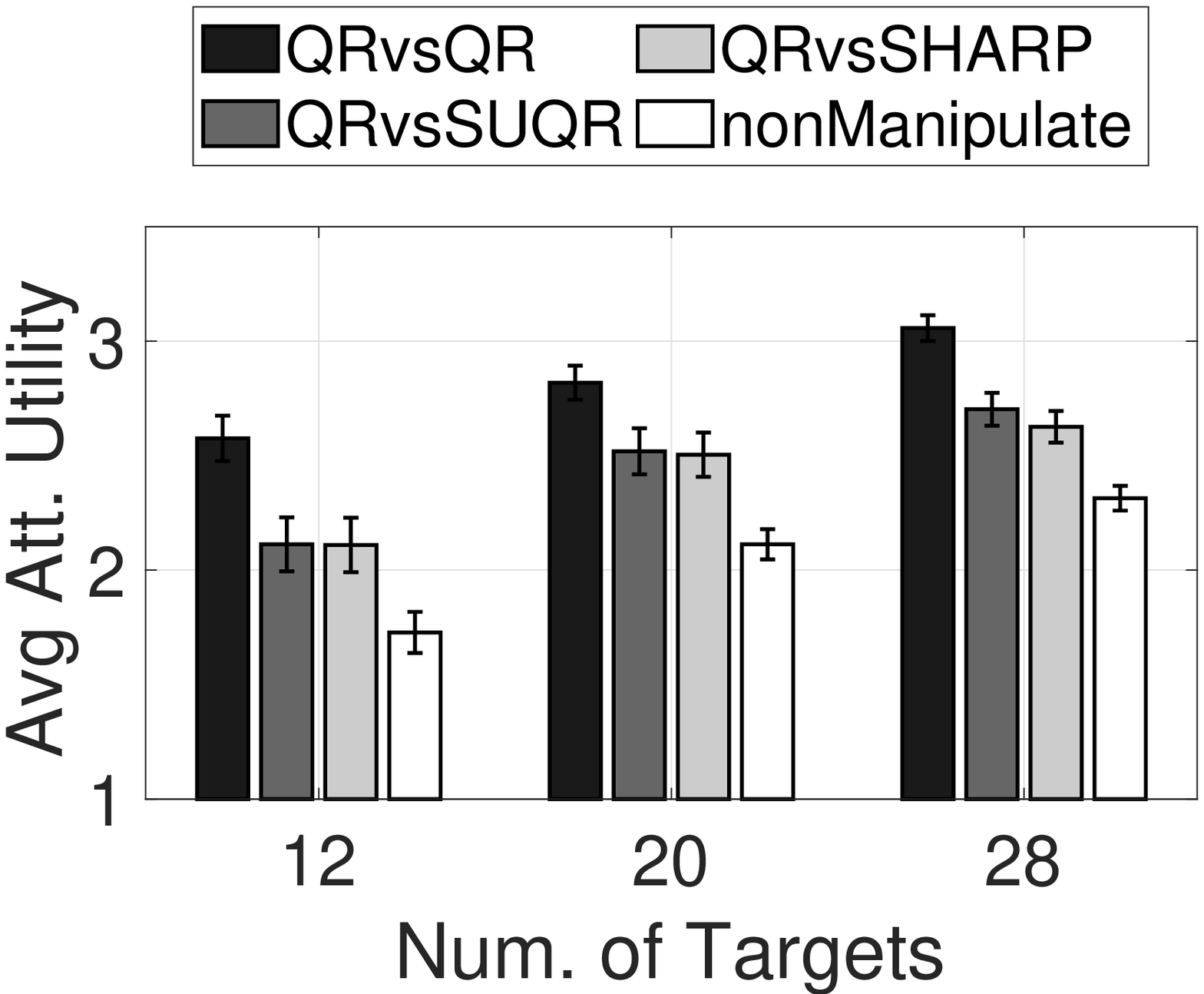}}
    \subfigure[QR manipulation, $T=8$]{\includegraphics[width = 0.33\textwidth]{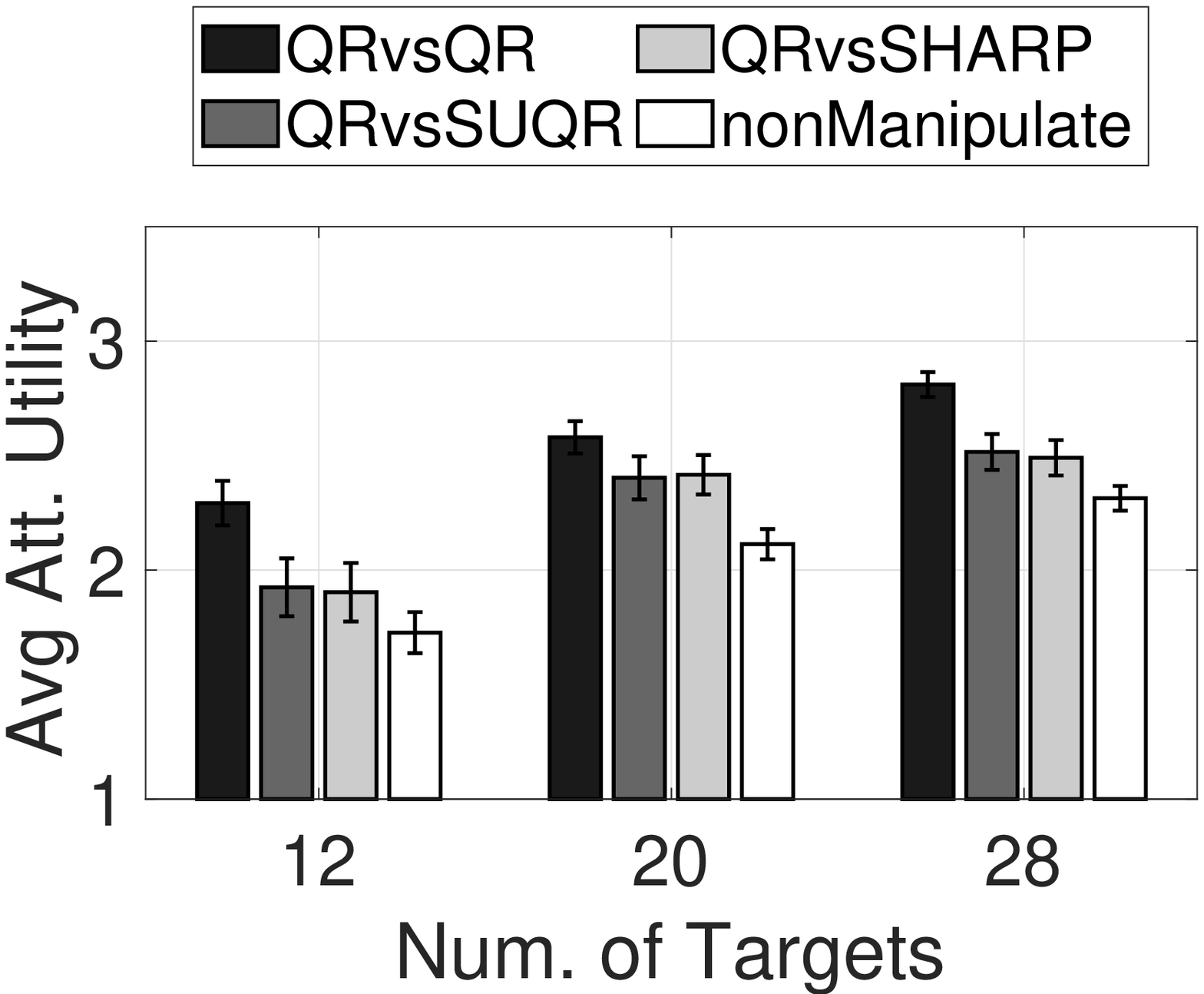}}
    \subfigure[SUQR manipulation, $T=4$]{\includegraphics[width = 0.33\textwidth]{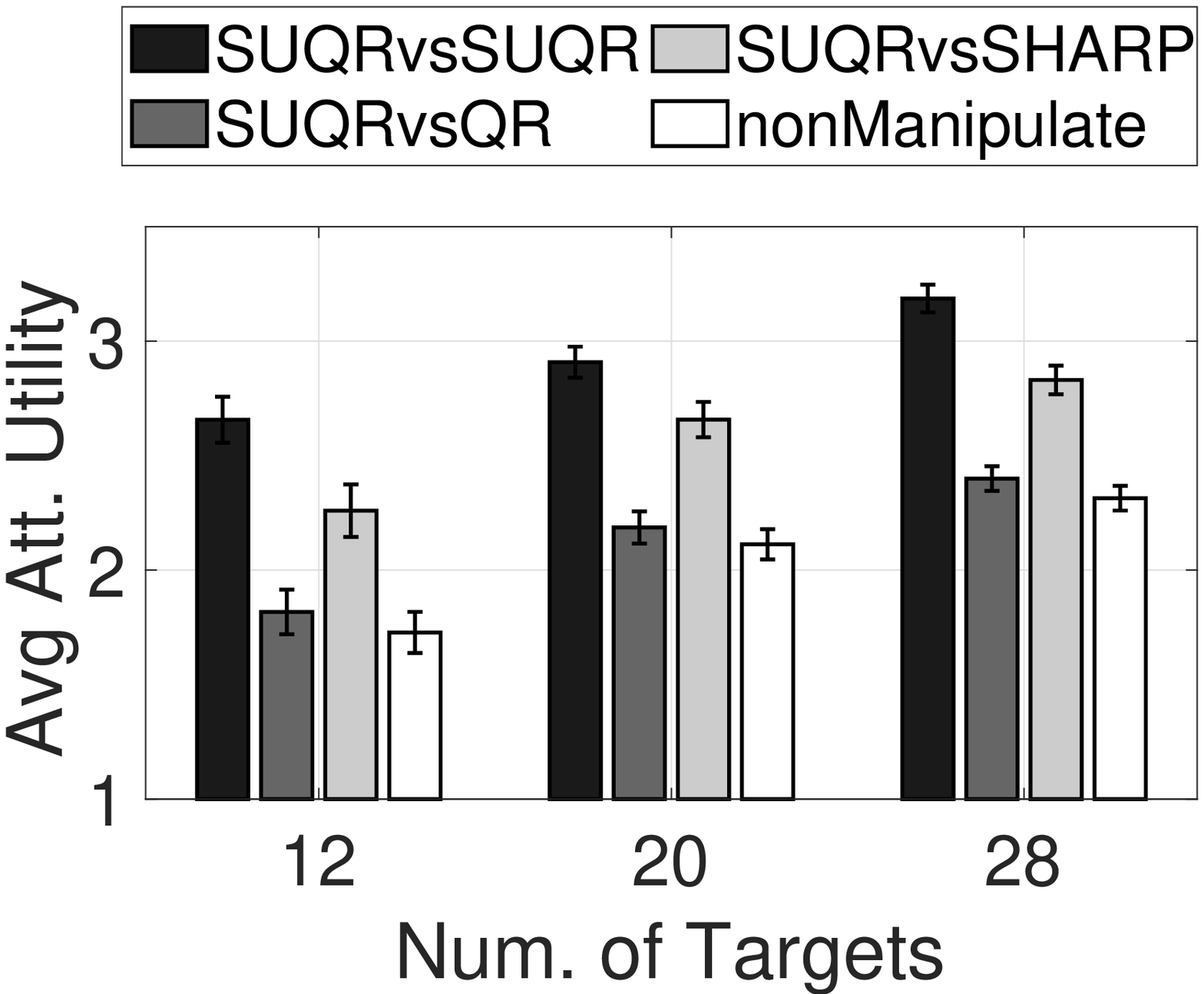}}
    \subfigure[SUQR manipulation, $T=8$]{\includegraphics[width = 0.33\textwidth]{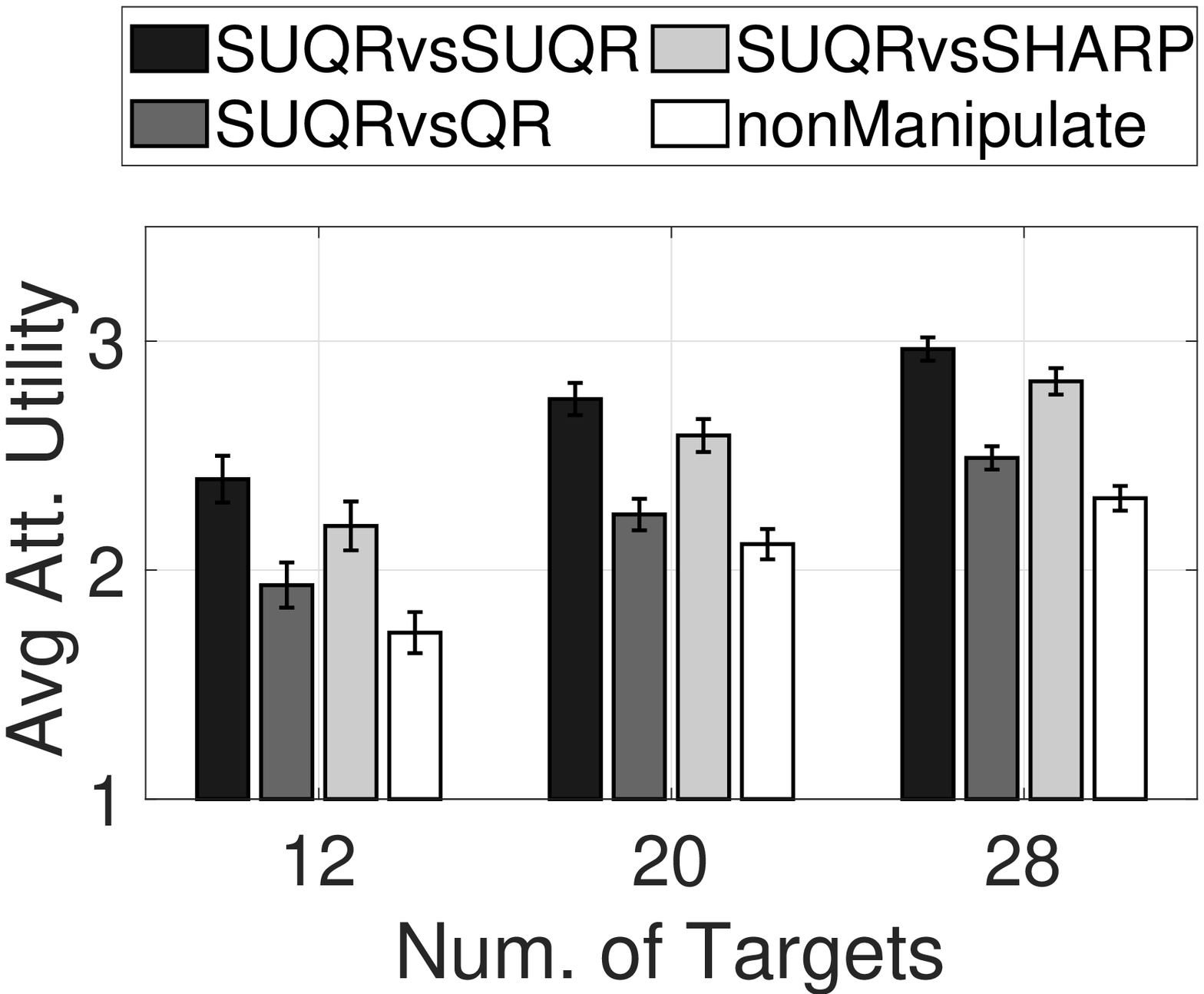}}
    \subfigure[SHARP manipulation, $T\!=\!4$]{\includegraphics[width = 0.33\textwidth]{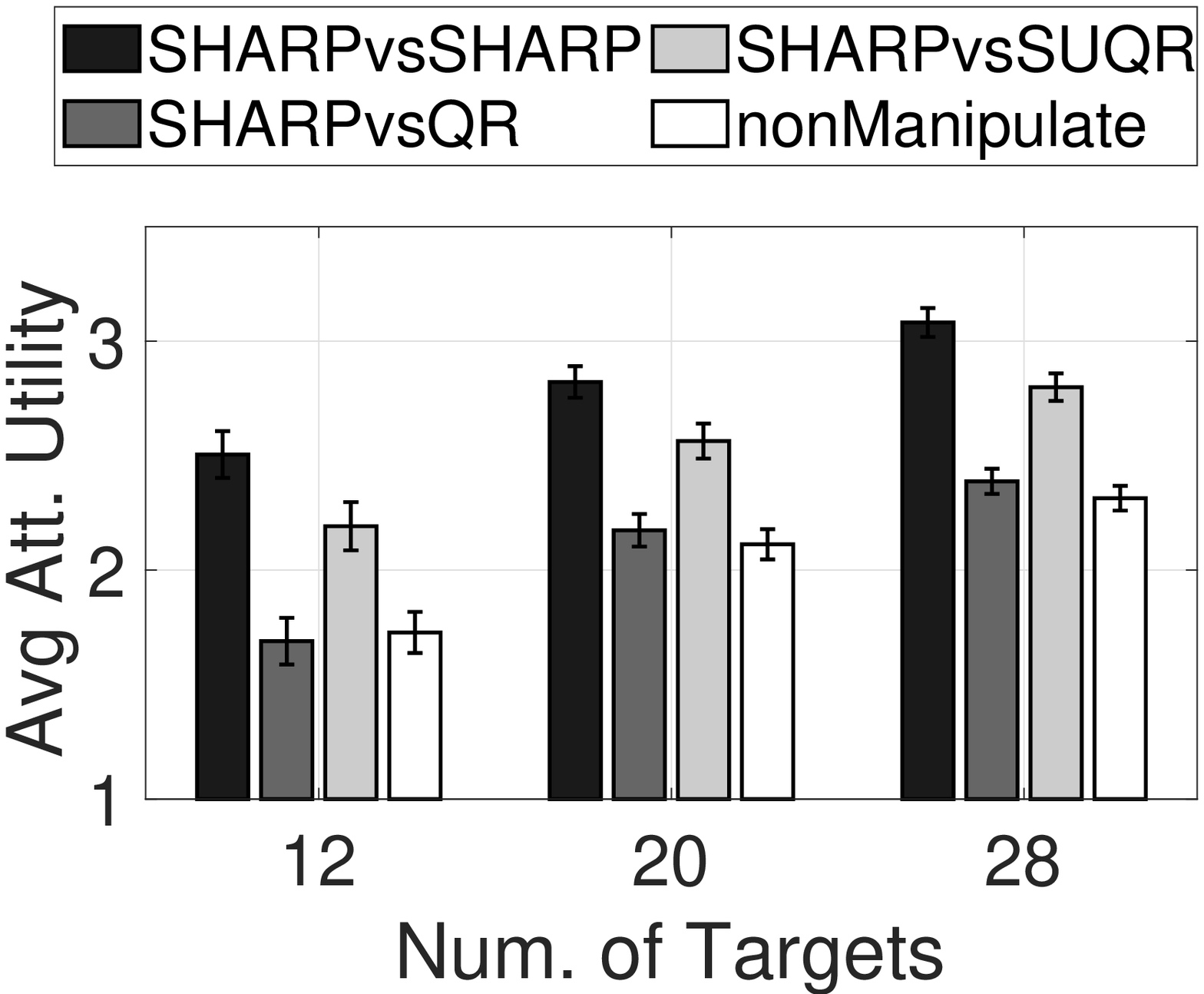}}
    \subfigure[SHARP manipulation, $T\!=\!8$]{\includegraphics[width = 0.33\textwidth]{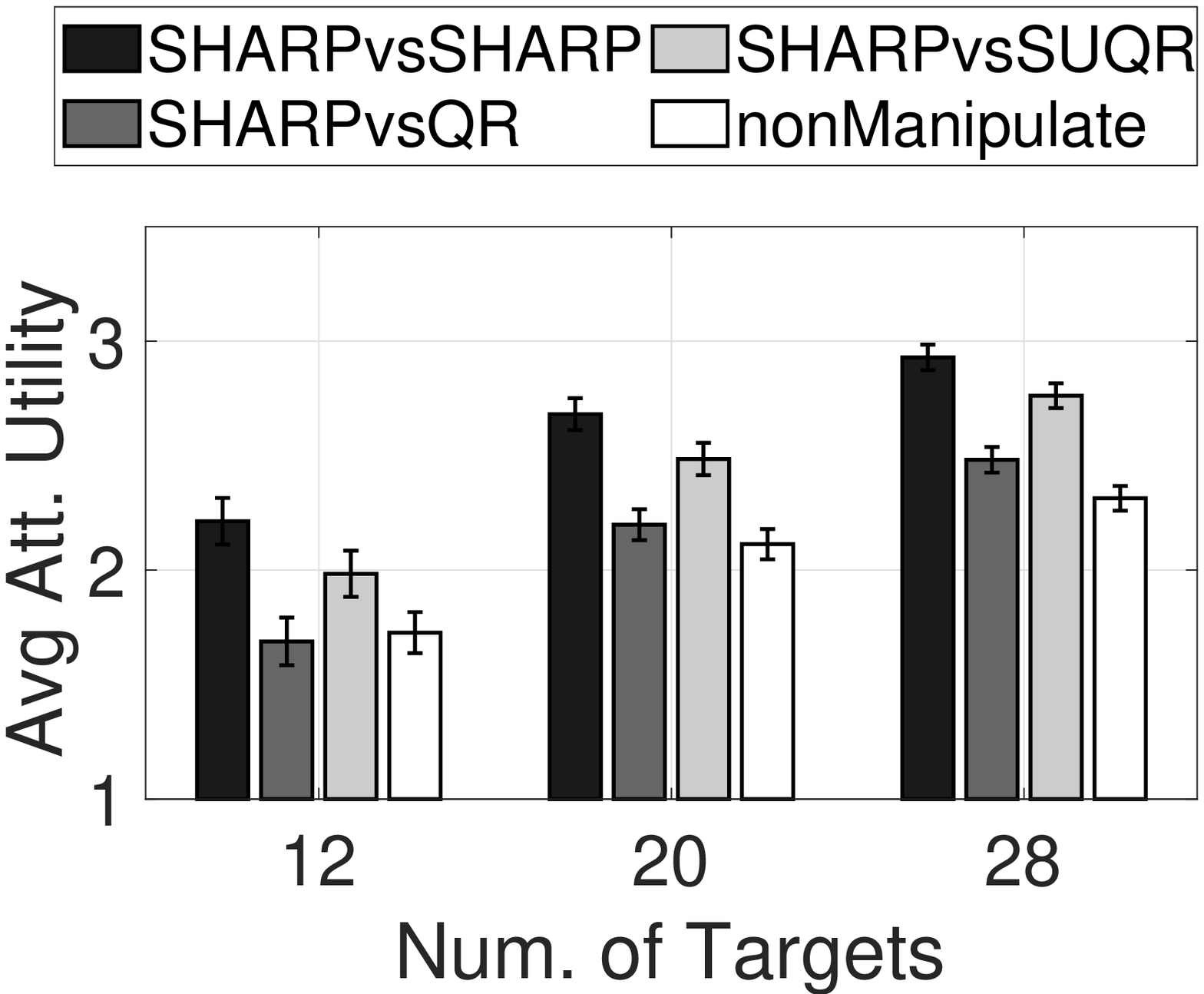}}
    \caption{Attacker Utility Evaluation}
    \label{fig:attU0.3R}
\end{figure}

\begin{figure}[t!]
    \centering
    \subfigure[QR manipulation, $T=4$]{\includegraphics[width = 0.33\textwidth]{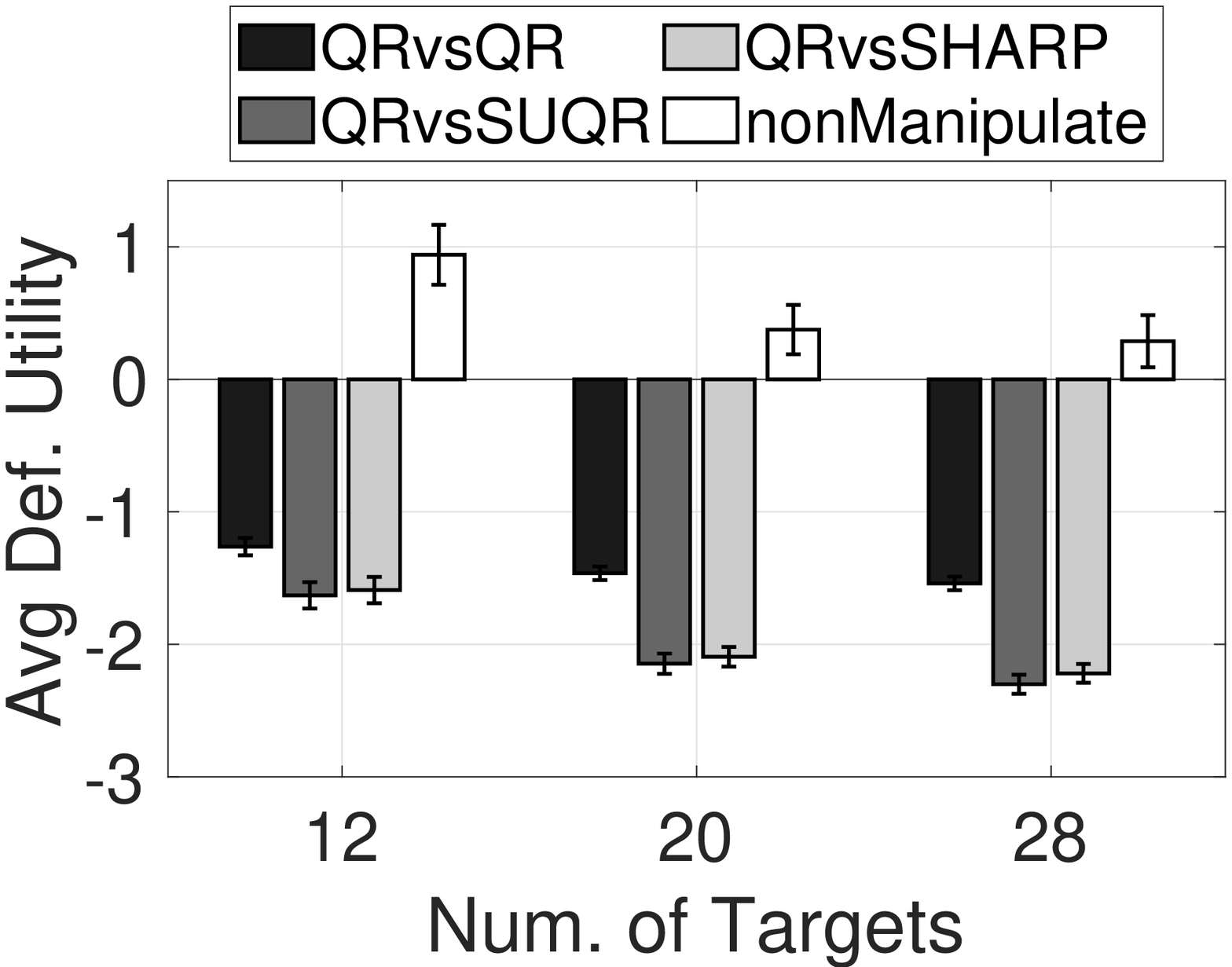}}
    \subfigure[QR manipulation, $T=8$]{\includegraphics[width = 0.33\textwidth]{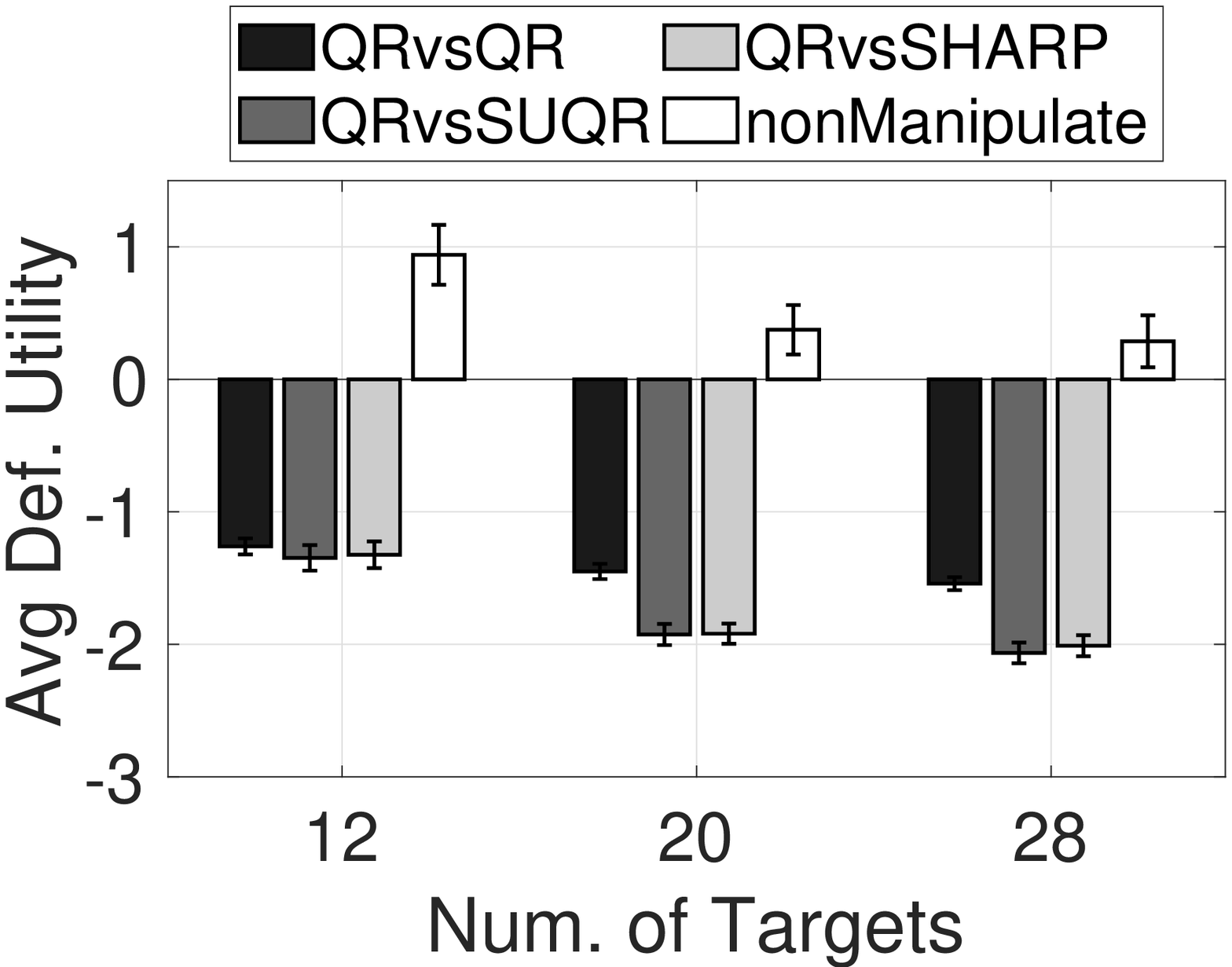}}
    \subfigure[SUQR manipulation, $T=4$]{\includegraphics[width = 0.33\textwidth]{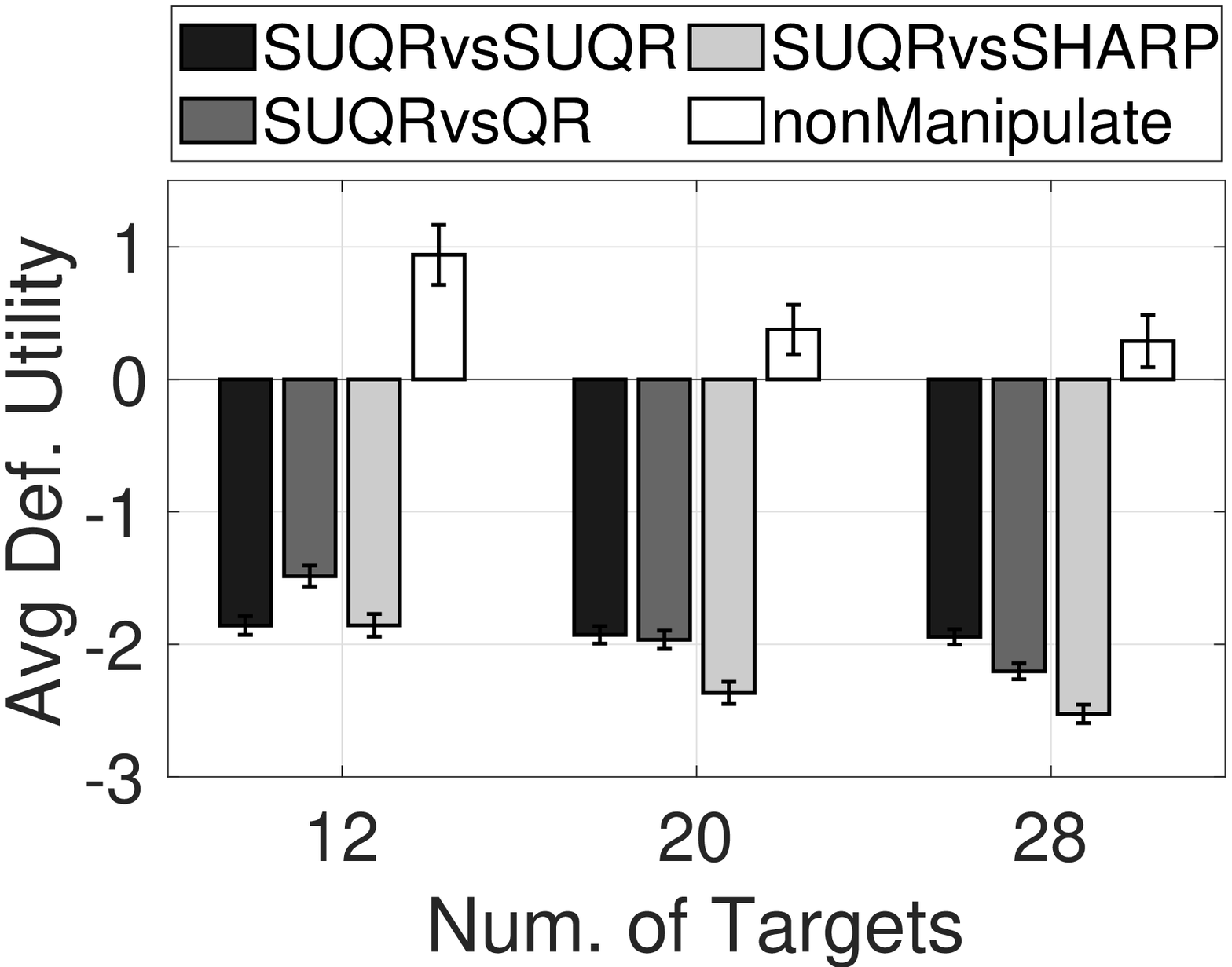}}
    \subfigure[SUQR manipulation, $T=8$]{\includegraphics[width = 0.33\textwidth]{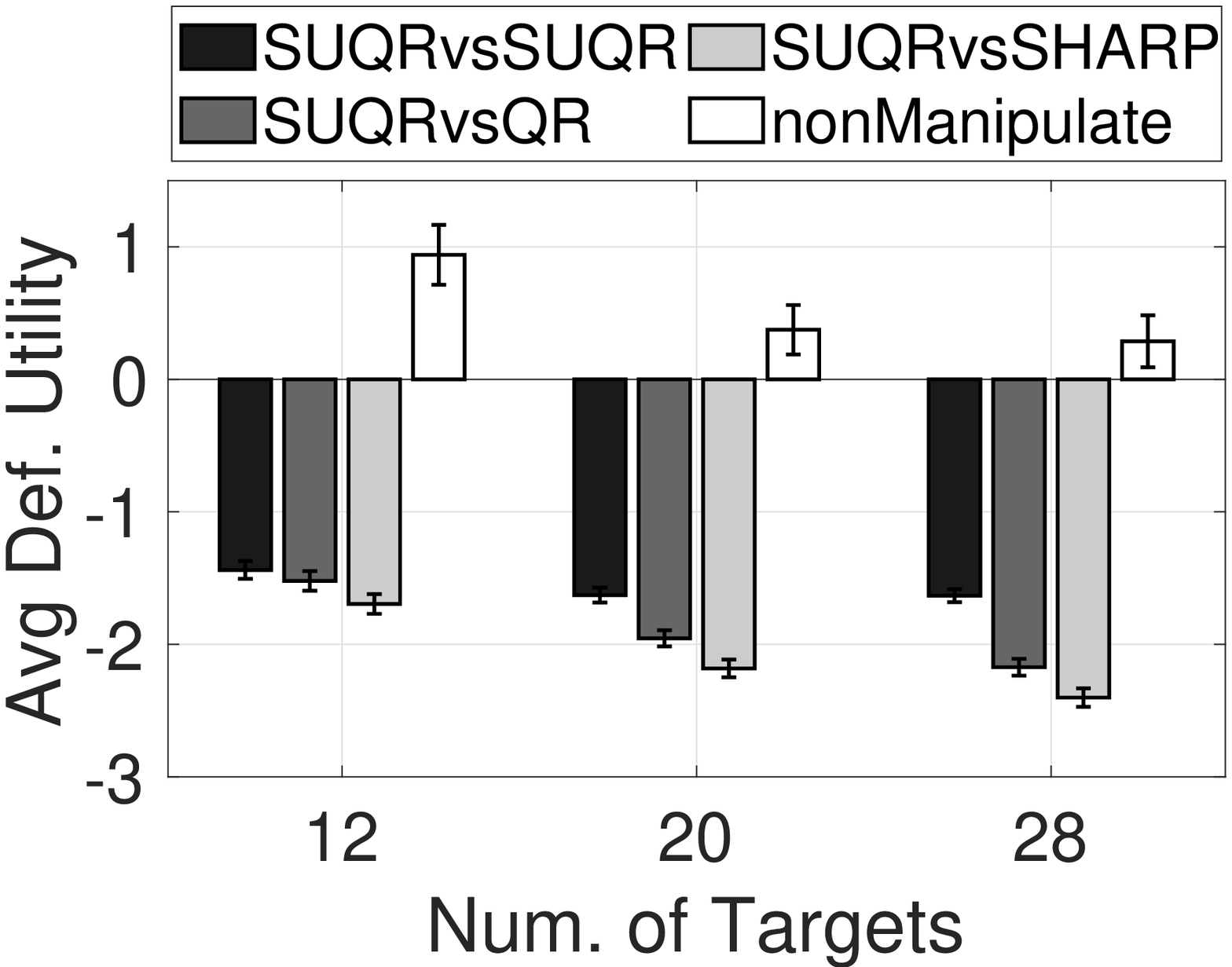}}
    \subfigure[SHARP manipulation, $T\!=\!4$]{\includegraphics[width = 0.33\textwidth]{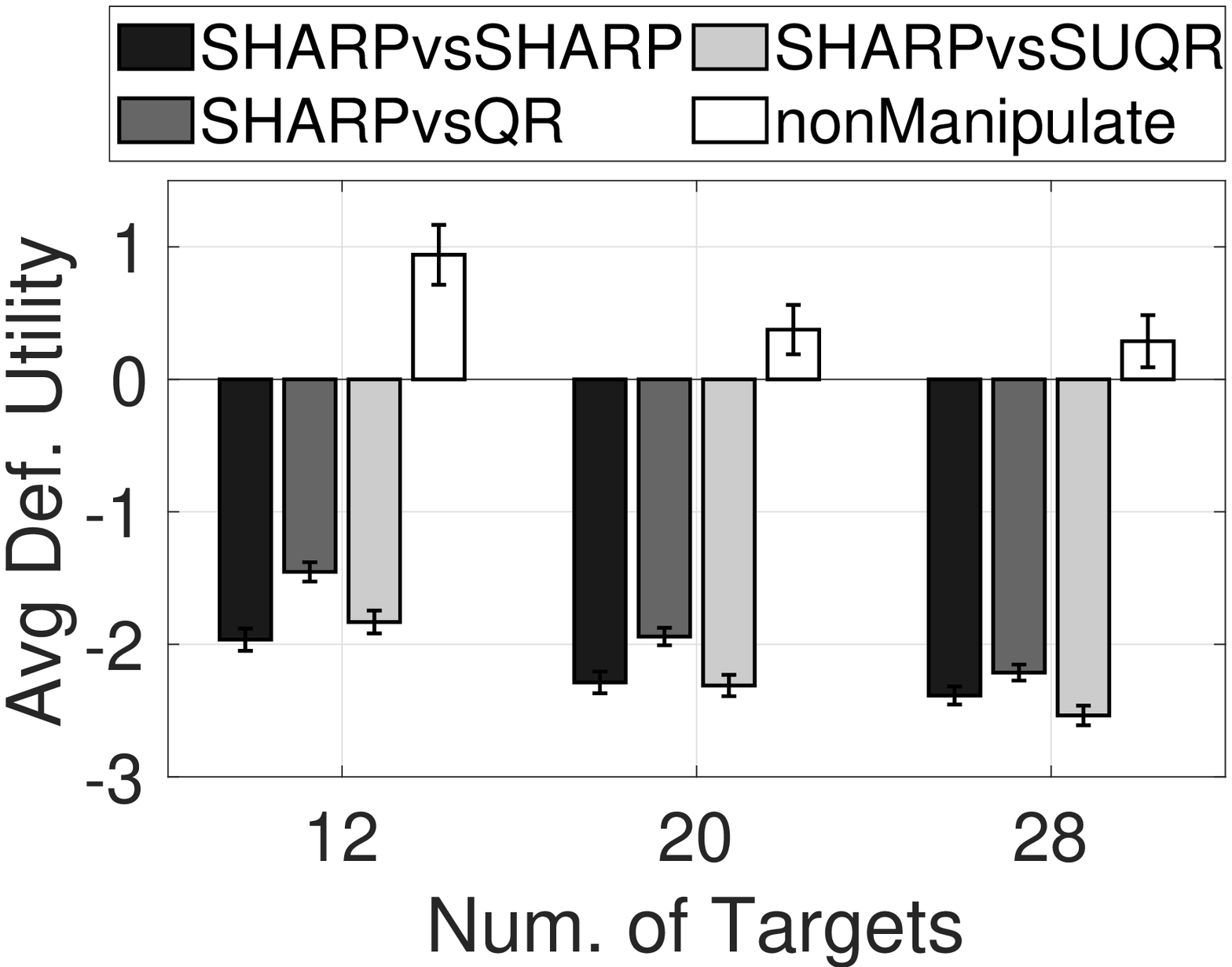}}
    \subfigure[SHARP manipulation, $T\!=\!8$]{\includegraphics[width = 0.33\textwidth]{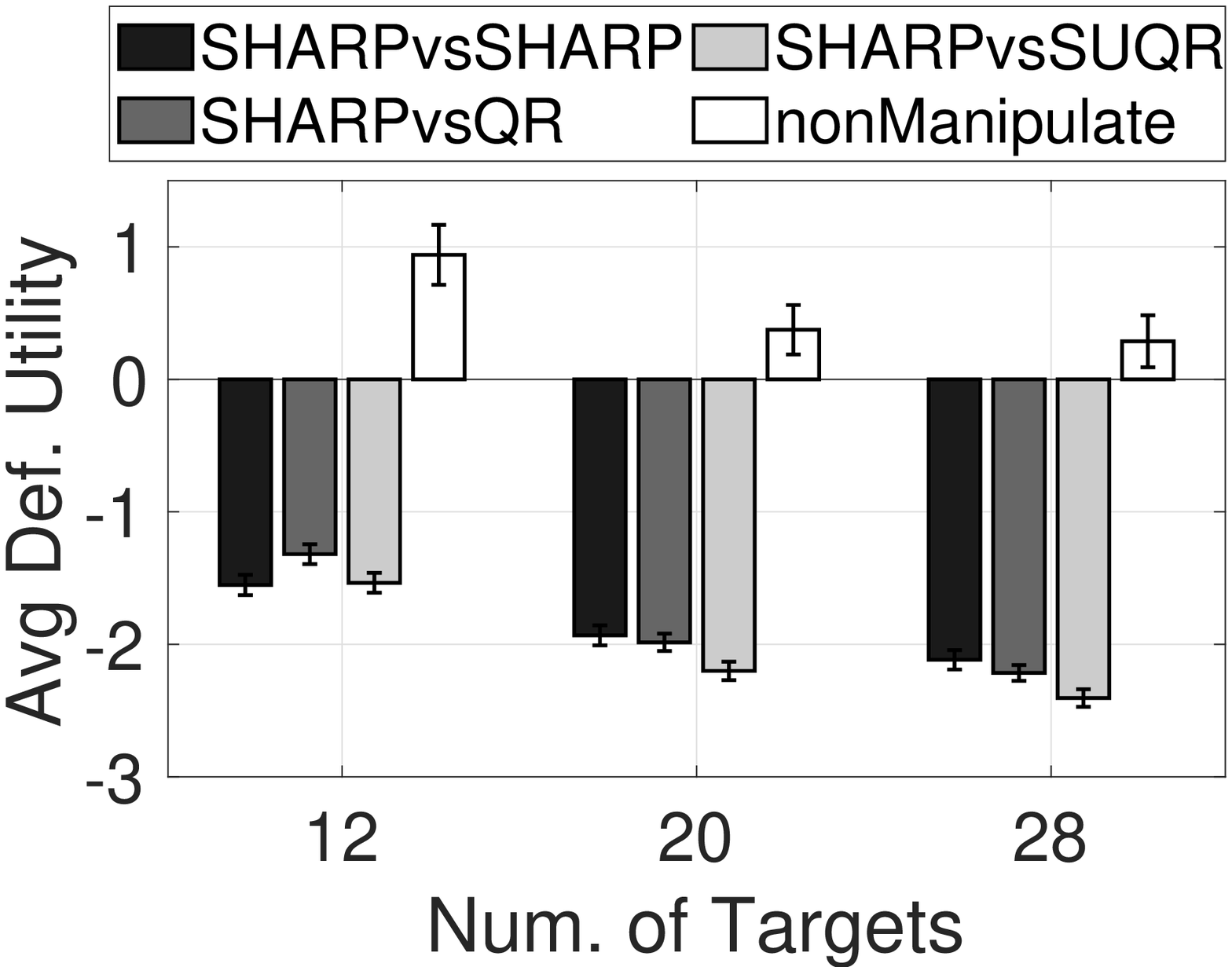}}
    \caption{Defender Utility Evaluation}
    \label{fig:defU0.3R}
\end{figure}

In the main paper, we present our experimental results in the game settings in which the defender can only protect up to 50\% of the targets (i.e., the number of security resources is equal to 50\% of the number of targets). In addition to these experiments, we evaluate the attacker manipulation in the case when the resource-target ratio is 30\%. The results are shown in Figures~(\ref{fig:attU0.3R}--\ref{fig:defU0.3R}). Overall, we see similar trends in these figures. However, in terms of the attacker utility (Figure~\ref{fig:attU0.3R}), the impact of attacker's manipulation on the defender's strategies are less than when the ratio is 50\% (Figure~\ref{fig:attU}). This result make sense because when the strategy space of the defender is smaller (i.e., the resource-target ratio is 30\%), there is less chance for the attacker to manipulate the defender's selection of strategies. In addition, given a smaller security resource-target ratio, the immediate expected utility of the attacker for playing the myopic best response (i.e., non-manipulative response) at each time step is higher. This is because the attacker's immediate expected utility at each target is a decreasing function of the defender's coverage probability at that target. As a result, the attacker has less incentive to deviate from that myopic best response. In other words, the attacker is less likely to pretend to be boundedly rational.  
\end{document}